\documentclass{article} % For LaTeX2e
\usepackage{iclr2023_conference,times}

\usepackage{amsmath,amsfonts,bm}

\def\eqref#1{equation~\ref{#1}}
\def\1{\bm{1}}

\DeclareMathAlphabet{\mathsfit}{\encodingdefault}{\sfdefault}{m}{sl}
\SetMathAlphabet{\mathsfit}{bold}{\encodingdefault}{\sfdefault}{bx}{n}

\usepackage{url}

\usepackage{makecell}
\usepackage{multirow}
\usepackage{graphicx}
\usepackage{amsmath}
\usepackage{amssymb}
\usepackage{booktabs}
\usepackage{subfig}
\usepackage{wrapfig}
\usepackage[colorlinks,
            linkcolor=red,
            anchorcolor=blue,
            citecolor=green,
            ]{hyperref}

\title{PatchDCT: Patch Refinement for High Quality Instance Segmentation}

\author{Qinrou Wen$^{1}$, Jirui Yang$^{2}$, Xue Yang$^{3}$, Kewei Liang$^{1,}$\thanks{Corresponding author is Kewei Liang.}\\
$^{1}$School of Mathematical Sciences, Zhejiang University \quad $^{2}$Alibaba Group\\ 
$^{3}$Department of CSE, MoE Key Lab of Artificial Intelligence, Shanghai Jiao Tong University \\
\texttt{\{qinrou.wen,matlkw\}@zju.edu.cn, \quad jirui.yjr@alibaba-inc.com}\\
\texttt{yangxue-2019-sjtu@sjtu.edu.cn}\\
\texttt{PyTorch Code: \url{https://github.com/olivia-w12/PatchDCT}}
}
\iclrfinalcopy % Uncomment for camera-ready version, but NOT for submission.
\begin{document}

\maketitle

\newcommand{\yangr}[1]{{\color{black} #1}}
\newcommand{\wqr}[1]{{\color{black} #1}}
\begin{abstract}
High-quality instance segmentation has shown emerging importance in computer vision. Without any refinement, DCT-Mask directly generates high-resolution masks by compressed vectors. 
To further refine masks obtained by compressed vectors, we propose for the first time a compressed vector based multi-stage refinement framework.  However, the vanilla combination does not bring significant gains, because changes in some elements of the DCT vector will affect the prediction of the entire mask. Thus, we propose a simple and novel method named PatchDCT, which separates the mask decoded from a DCT vector into several patches and refines each patch by the designed classifier and regressor.
Specifically, the classifier is used to distinguish mixed patches from all patches, and to correct previously mispredicted foreground and background patches. In contrast, the regressor is used for DCT vector prediction of mixed patches, further refining the segmentation quality at boundary locations. Experiments on COCO show that our method achieves 2.0\%, 3.2\%, 4.5\% AP and 3.4\%, 5.3\%, 7.0\% Boundary AP improvements over Mask-RCNN on COCO, LVIS, and Cityscapes, respectively. It also surpasses DCT-Mask by 0.7\%, 1.1\%, 1.3\% AP and 0.9\%, 1.7\%, 4.2\% Boundary AP on COCO, LVIS and Cityscapes. Besides, the performance of PatchDCT is also competitive with other state-of-the-art methods.
%, and the code will be made publicly available.
\end{abstract}

\section{Introduction}\label{sec:intro}
Instance segmentation  \citep{li2017fully,he2017mask} is a fundamental but challenging task in computer vision, which aims to locate objects in images and precisely segment each instance. 
% Instance segmentaion methods, which follows mask-rcnn paradigm, often segment instance in a low resolution grid~\cite{kang2020bshapenet,cheng2020boundary}.
The mainstream instance segmentation methods follow Mask-RCNN  \citep{he2017mask} paradigm, which often segment instances in a low-resolution grid \citep{kang2020bshapenet,cheng2020boundary,chen2019hybrid, ke2021deep}.
However, limited by the coarse mask representation ( i.e. $28\times 28$ in Mask-RCNN), most of these algorithms cannot obtain high-quality segmentation results due to the loss of details. DCT-Mask  \citep{shen2021dct} achieves considerable performance gain by predicting an informative $300$-dimensional Discrete Cosine Transform (DCT) \citep{ahmed1974discrete} vector compressed from a $128\times 128$ mask.
%In order to generate masks with higher quality, it's a natural idea to further improve the segmentation results of DCT-Mask, 
To further improve the segmentation results of DCT-Mask, we follow the refine mechanism  \citep{ke2022mask, zhang2021refinemask, kirillov2020pointrend} to correct the mask details in a multi-stage manner.

\begin{figure}[!tb]
    \centering
	\subfloat[Influence of element change in DCT-Mask]{
		\label{fig1:dct_changes}
		\includegraphics[width=0.48\textwidth]{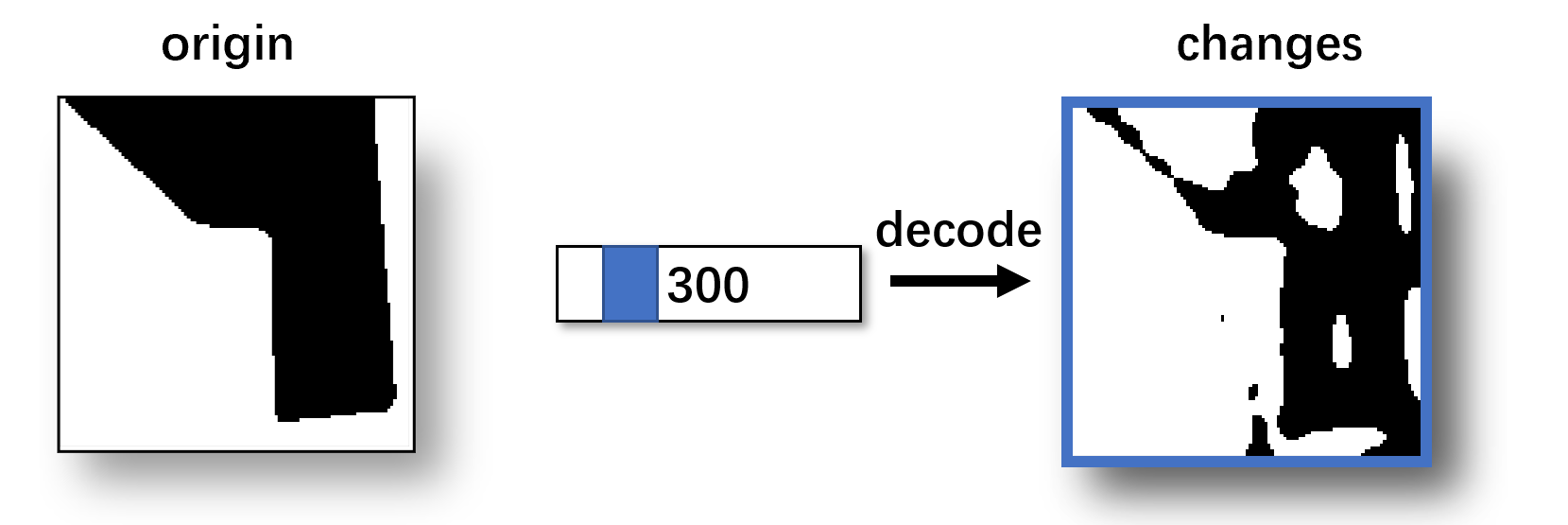}
	}
	\subfloat[Influence of element change in PatchDCT]{
		\label{fig1:patch_changes}
		\includegraphics[width=0.48\textwidth]{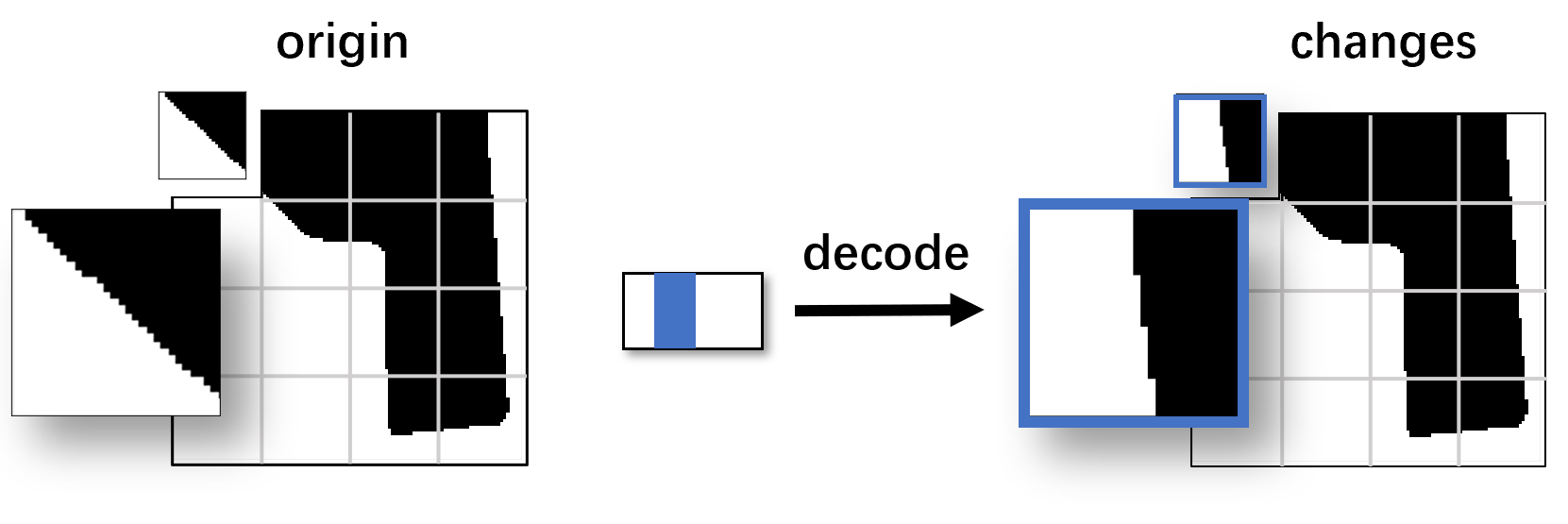}
	}
	% \vspace{2mm}
	%\vspace{-2mm}
	\caption{(a) Influence of elements changes in DCT vectors for DCT-Mask. The blue block denotes the changed elements. The box with a blue border represents the part of the mask affected by the changes in element values. The change of some elements will affect the entire mask. (b) Influence of elements changes in DCT vectors for PatchDCT. Changing some elements of a vector will only affect the corresponding patch.}
	\label{fig: influence}
	\vspace{-4mm}
\end{figure}

A straightforward implementation is to refine the $300$-dimensional DCT vector multiple times. 
However, experimental results show that this naive implementation does not succeed, 
which improves mask average precision (mAP) by 0.1\% from 36.5\% to 36.6\% on COCO \emph{val set}.
The main reason for the limited improvement is that the full $300$-dimensional DCT vector is not suitable for refining some important local regions, such as wrong predicted regions and boundary regions in masks. 
%Referring to Figure \ref{fig1:dct_changes}, changes in some elements of the DCT vector will affect the prediction of the entire mask,  including the correctly segmented regions.
%As each pixel value in the mask is calculated by all elements of the DCT vector in the inference stage, once the predicted value of some elements in the DCT vector changes, the entire mask will change, and even the correct segmentation area may be affected, refer to Figure \ref{fig1:dct_changes}.
%As each pixel value in the mask is calculated by all elements of the DCT vector in the inference stage, changing some elements of the DCT vector will affect the prediction of the entire mask,  including the correctly segmented regions, refer to Figure \ref{fig1:dct_changes}.
As each pixel value in the mask is calculated by all elements of the DCT vector in the inference stage, once some elements in the DCT vector change, the entire mask will change, and even the correct segmentation areas may be affected, refer to Figure \ref{fig1:dct_changes}.

To overcome the above issue, we propose a novel method, called PatchDCT, which divides the mask decoded from a DCT vector into several independent patches and refines each patch with a three-class classifier and a regressor, respectively. In detail, each patch is first classified into one of three categories: foreground, background, and mixed by the classifier, and then previously mispredicted foreground and background patches will be corrected. %In contrast, 
Mixed patches are fed into the regressor to predict their corresponding $n$-dimensional ($n \ll 300$) DCT vectors. In the inference stage, we use  Inverse Discrete Cosine Transform (IDCT) to decode the predicted vectors of the mixed patches as their refined masks, and merge them with the masks of other foreground and background patches to obtain a high-resolution mask.
%\yangr{Compared with DCT-Mask, PatchDCT does not introduce too much computation and parameters even if multiple stages are used, and each refinement stage only needs to predict the category of each patch and the DCT vector of mixed patches.} 
It is also worth emphasizing that each patch is independent, so the element change of a DCT vector will only affect the corresponding mixed patch, as shown in Figure \ref{fig1:patch_changes}. %Figure \ref{fig1:patch_changes} shows the influence of element changes in a DCT vector on the prediction results of PatchDCT. 
In general, patching allows the model to focus on the refinement of local regions, thereby continuously improving the quality of segmentation, resulting in significant performance improvements. Our main contributions are:

% To address these problems, we propose a method named PatchDCT. PatchDCT contains a three-class classifier and a regressor. After obtaining a high-resolution mask by a DCT vector, we divide the mask into several patches and refine each patch respectively. The patches that have only foreground pixels, only background pixels, both foreground and background pixels are named "foreground", "background" and "mixed" patches respectively. By distinguishing the patches, the three-class classifier correct the previously mispredicted foreground and background patches and identifies the boundaries (namely the mixed patches) of masks. The regressor further refines boundaries of masks by predicting a $n$-dimensional DCT vector ($n\ll 300$) for each mixed patch. In inference time, we decode the vectors and combine all the patches of a mask to generate a high-resolution mask (Figure~\ref{fig1:inference_patch}). PatchDCT allows a low training complexity by only predicting the category of each patch and the DCT vectors of mixed patches. In PatchDCT, each classification result corresponds to a patch, and the changes of an element of a DCT vector will only influence the corresponding patch (Figure~\ref{fig1:patch changes}). In other words, the refinement of each patch then does not influence each other.

\textbf{1)} To our best knowledge, PatchDCT is the first compressed vector based multi-stage refinement detector to predict high-quality masks.

\textbf{2)} PatchDCT innovatively adopts the patching technique, which successfully allows the model to focus on the refinement of important local regions, fully exploiting the advantages of multi-stage refinement and high-resolution information compression.

\textbf{3)} Compared to Mask RCNN, PatchDCT improves about 2.0\% AP and 3.4\% Boundary AP on COCO, 3.2\% AP and 5.3\% Boundary AP on LVIS$^*$\footnote{COCO dataset with LVIS annotations}, 4.5\% AP and 7.0\% Boundary AP on Cityscapes. It also achieves 0.7\% AP and 0.9\% Boundary AP on COCO, 1.1\% AP and 1.7\% Boundary AP on LVIS$^*$, 1.3\% AP and  4.2\% Boundary AP on Cityscapes over DCT-Mask.

\textbf{4)} Demonstrated by experiments on COCO \emph{test-dev}, the performance of PatchDCT is also competitive with other state-of-the-art methods.
%, and the code will be made publicly available.

% PatchDCT is a refinement module that inputs a high-resolution mask and outputs a finer high-resolution mask. Therefore, the mask generated by PatchDCT can be feed into PatchDCT again to obtain a finer and finer mask. 
%We name the model that refines the mask $n$ times with PatchDCT "$n$-stage PatchDCT". 
% We evaluate PatchDCT on COCO, LVIS, and Cityscapes. In comparison with Mask-RCNN, PatchDCT improves about 2.2\% mAP and 3.5\% Boundary AP on COCO, 3.6\% mAP on LVIS, 4.5\% mAP and 7.0\% Boundary AP on Cityscapes . 
% It also achieves 0.9\% mAP and 1.1 Boundary AP on COCO, 1.8\% mAP on LVIS, 1.3\% mAP and  4.2\% Boundary AP on Cityscapes over DCT-Mask.
% Demonstrated by experiments on COCO \emph{test-dev2017}, the performance of PatchDCT is also competitive with other state-of-art methods.

%-------------------------------------------------------------------------
\section{Related Work}
%\textbf{Revisit instance segmentation from coarse to fine.}
\textbf{Instance segmentation.}
Instance segmentation assigns a pixel-level mask to each instance of interest. 
Mask-RCNN \citep{he2017mask} generates bounding boxes for each instance with a powerful detector \citep{ren2015faster} 
and categorizes each pixel in bounding boxes as foreground or background to obtain $28\times 28$ binary grid masks. 
Several methods that build on Mask-RCNN improve the quality of masks. %Cascade RCNN \citep{cai2018cascade} uses the same mask branch as Mask-RCNN with a multi-stage detector. 
Mask Scoring RCNN \citep{huang2019mask} learns to regress mask IoU to select better-quality instance masks.
%instance masks with better quality. 
HTC \citep{chen2019hybrid} utilizes interleaved execution, mask information flow, and semantic feature fusion to improve Mask-RCNN. BMask RCNN \citep{cheng2020boundary} adds a boundary branch on Mask-RCNN to detect the boundaries of masks. Bounding Shape Mask R-CNN \citep{kang2020bshapenet} improves performance on object detection and instance segmentation by its bounding shape mask branch. BCNet \citep{ke2021deep} uses two GCN \citep{welling2016semi} layers to detect overlapping instances. 
%In addition to these two-stage based algorithms, some one-stage algorithms, such as PolarMask \citep{xie2020polarmask}, SOLO \citep{wang2020solo,wang2020solov2} , etc, have been developed. 
Although these algorithms have yielded promising results, they are still restricted 
in the low-resolution mask representation and thus do not generate high-quality masks.
%Some methods explore using other representations of masks besides binary grid to improve segmentation. PolarMask\cite{xie2020polarmask} regards mask prediction as  contour prediction in the polar coordinates. 

\textbf{Towards high-quality instance segmentation.} 
%Since many details may be lost by low-resolution masks, recent methods concentrate on generating fine high-resolution masks. 
To take full advantage of high-resolution masks, DCT-Mask \citep{shen2021dct} learns to regress a $300$-dimensional DCT vector compressed from a $128\times 128$ mask. SOLQ  \citep{dong2021solq} is a query-based method, which also encodes high-resolution masks into DCT vectors and predicts the vectors by queries. Both of these methods generate high-resolution masks in a one-shot manner, without any refinement. %Although considerable gains be made, 
Although they have made considerable gains, there is still potential for improvement. Multi-stage refinement is another common technique for obtaining high-quality masks.
%To generate high-quality mask, a popular technique is refining a coarse mask multiple times to obtain a high-resolution masks.
%PointRend  \citep{kirillov2020pointrend} adaptively selects several locations to compute labels and applies an iterative subdivision algorithm to generate $224\times 224$ masks from $7\times 7$ coarse masks. 
PointRend \citep{kirillov2020pointrend} adaptively selects several locations to refine, %and generates
rendering $224\times 224$ masks from $7\times 7$ coarse masks.
%RefineMask  \citep{zhang2021refinemask} leverage multi-stage procedure and boundary-aware refinement to derive $112\times 112$ masks from $14\times 14$ masks.
RefineMask  \citep{zhang2021refinemask} introduces semantic segmentation masks as auxiliary inputs, and generates $112\times 112$ masks in a multi-stage manner.
Mask Transfiner  \citep{ke2022mask} represents image regions as a quadtree and corrects the errors of error-prone tree nodes to generate $112\times 112$ masks. PBR \citep{tang2021look} is a post-processing method that refines patches along the mask boundaries. 
%Unlike these methods, which refine on binary grid masks, our refinement method is based on compressed vectors.
Unlike these refinement methods based on the binary grid mask representation, our method is based on compressed vectors.
%which are based on binary-grid mask for refinement, our method refines the compressed vector.

\wqr{Generating high-quality masks is also one of the main concerns in the field of semantic segmentation. 
%CRF \citep{krahenbuhl2011efficient} is used in many works of semantic segmentation to improve the performance \citep{zheng2015conditional,chen2017deeplab,shen2017semantic}.
CRFasRNN \citep{zheng2015conditional} connects CRF \citep{krahenbuhl2011efficient} with FCN \citep{long2015fully}, formulating mean-field approximate inference for the CRF with Gaussian pairwise potentials as Recurrent Neural Networks.
%On the basis of deep convolutional neural networks, 
DeepLab \citep{chen2017deeplab} effectively improves the quality of masks by using atrous convolution for receptive field enhancement, ASPP for multi-scale segmentation, and CRF for boundary refinement.
SegModel \citep{shen2017semantic} utilizes a guidance CRF to improve the segmentation quality. CascadePSP \citep{cheng2020cascadepsp} trains independently a refinement module designed in a cascade fashion. %In addition to CRF, other post-processing methods are also used in semantic segmentation. CascadePSP \citep{cheng2020cascadepsp} trains independently an refinement module designed in a cascade fashion.
RGR \citep{dias2018semantic} is a post-processing module based on region growing. In contrast, PatchDCT can obtain high-quality segmentation results in an end-to-end learning manner without any additional post-processing.

% These methods usually rely on post-processing modules to refine boundary segmentation results, while PatchDCT is learned in an end-to-end manner.
%Compared to these post-processing module that need to be trained independently, PatchDCT is a refinement module connected with DCT-Mask for end-to-end training.

}

\begin{figure}
	\centering
	\includegraphics[width=1.0\textwidth]{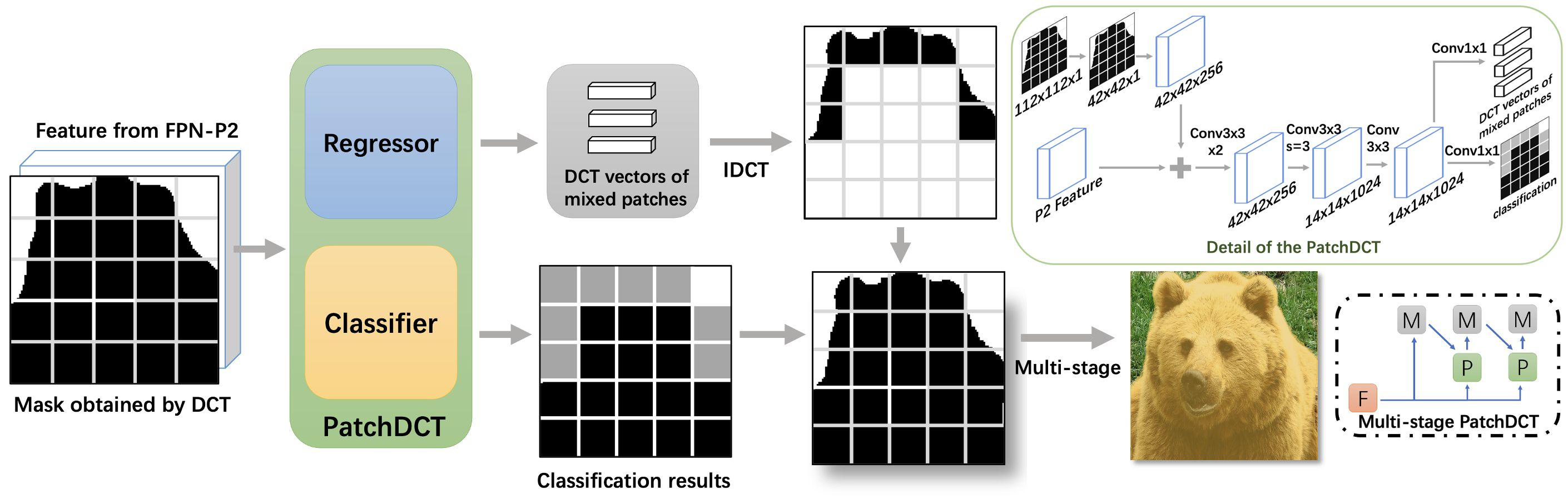}
	\caption{The pipeline of PatchDCT. The classifier differentiates foreground, background and mixed patches. The regressor predicts the DCT vectors of mixed patches. Masks of mixed patches are obtained by patch DCT vectors. PatchDCT combines masks of all patches to obtain an entire mask of instance. The entire mask of instance output by PatchDCT can be fed into another PatchDCT module for a finer mask.
	For the architecture of multi-stage PatchDCT:
	`F' is the feature map cropped from FPN-P2.
	`M' is the high-resolution mask.
	`P' is the PatchDCT module.}
	\label{fig:patchdctmodel}
 \vspace{-10pt}
\end{figure}

\section{Methods}
In this section, we show the difficulties in refining DCT vectors and then introduce PatchDCT to overcome these difficulties and generate finer masks.
\subsection{Difficulties in Refining DCT Vectors}\label{sec:diff}
Given a $K\times K$ mask, DCT-Mask \citep{shen2021dct} encodes the mask $\textbf{M}_{K\times K}$ into 
the frequency domain $\textbf{M}^f_{K\times K}$:
\begin{align}
	M^f_{K\times K}(u,v) = \frac{2}{K}C(u)C(v)\sum_{x=0}^{K-1}\sum_{y=0}^{K-1}
	M_{K\times K}(x,y) \cos\frac{(2x+1)u\pi}{2K}cos\frac{(2y+1)v\pi}{2K} ,
	\label{eq: dct}
\end{align}
where $C(w) = 1/\sqrt{2}$ for $w=0$ and $C(w)=1$ otherwise. 
Non-zero values are concentrated in the upper left corner of $\textbf{M}^f_{K\times K}$, 
which are low-frequency elements that contain the most information of the mask. 
%The high-frequency elements are located in the lower right corner of $\textbf{M}^f_{m\times m}$ and are almost all zeroes.
The $N$-dimensional DCT vector is obtained by zigzag scanning \citep{al2013jpeg} $\textbf{M}^f_{K\times K}$  and selecting the top-$N$ elements.

In the inference stage, $\textbf{M}^f_{K\times K}$ is recovered by filling the remaining elements to zero. Then each pixel in the mask $\textbf{M}_{K\times K}$ is calculated as follow:
\begin{align}
	M_{K\times K}(x,y) = \frac{2}{K}C(x)C(y)\sum_{u=0}^{K-1}\sum_{v=0}^{K-1}
	M^f_{K\times K}(u,v) \cos\frac{(2x+1)u\pi}{2K}cos\frac{(2y+1)v\pi}{2K},
	\label{eq: idct}
\end{align}

%DCT representation generates directly a high-resolution mask with low training complexity\cite{shen2021dct}.Further refinement of the mask will likely improve the quality of segmentation.However, it is not advisable to refine the high-resolution mask by reclassifying each pixel, since high training complexity of this method will cause bad prediction.Another idea is to predict a more accurate DCT vector for the mask.
%But there are still problems with this approach.

Equation \ref{eq: idct} reveals that each pixel in the mask  $\textbf{M}_{K\times K}$ is calculated by all elements of $\textbf{M}^f_{K\times K}$.
%When refining the $N$-dimensional DCT vector, it is nearly impossible to guarantee that all elements are correctly refined. Once an element is incorrectly changed, 
When refining the $N$-dimensional DCT vector, once an element is incorrectly changed, all pixels in $\textbf{M}_{K\times K}$ will be affected, even those correctly segmented regions, which is also shown in Figure \ref{fig: influence}.
Therefore, when fixing some specific error regions (e.g. borders), it is difficult to get the correct refinement result unless all the elements in the DCT vector are correctly refined. %In practice, however, it is almost impossible to correctly refine all $N$ elements.
In practice, however, it is almost impossible to correctly predict all $N$ elements.
%If we refine some elements of the DCT vector, every pixel value in $\textbf{M}_{K\times K}$ changes,
%The changes in each area may be positive or negative.
%so the correct segmentation area may be affect by the changes,which is also shown in Figure~\ref{fig: influence}.

%Thus, simply refining the DCT vector is unable to focus on the wrong segmentation areas and the areas that hard to predict.

\subsection{PatchDCT} \label{sec:patchdct}
To prevent the above issue when refining the global DCT vector, we propose a method named PatchDCT, which divides the $K\times K$ mask into $m\times m$ patches and refines each patch respectively.
The overall architecture of PatchDCT is shown in Figure \ref{fig:patchdctmodel}, which mainly consists of a three-class classifier and a DCT vector regressor.
Specifically, the classifier is used to identify mixed patches and refine foreground and background patches. 
Each mixed patch is then refined by an $n$-dimensional DCT vector, which is obtained from the DCT vector regressor.
%Mixed patches are refined by the DCT vector regressor, which regresses a $n$-dimensional DCT vector for each mixed patch.
%In the inference time, we obtain the class of each patch.Masks of foreground and background patches are directly obtained by the classifier.Masks for mixed patches are obtained by DCT vectors using Equation~\ref{eq: idct}.We then assemble all the patches to obtain a complete mask.The mask generated by PatchDCT can be further refined to obtain a finer mask.

\begin{wraptable}{r}{.45\textwidth}
% \begin{table}[tb!]
	\centering
 \caption{Mask AP obtained by different lengths of ground-truth DCT vectors using Mask-RCNN framework on COCO \emph{val2017}.
 The 1$\times$1 patch size represents the binary grid mask representation.
Low-dimensional DCT vectors are able to provide enough ground truth information.}
	% \scalebox{0.9}{
	\begin{tabular}{c|c|c|c}
		\Xhline{2\arrayrulewidth}
		\textbf{Resolution}&\textbf{Patch Size}&\textbf{Dim.}  & \textbf{AP}   \\ \hline
		$112\times 112$&$1\times 1$&1 & 57.6\\
		$112\times 112$&$8\times 8$&3 &55.8  \\ 
		$112\times 112$&$8\times 8$&6 &57.1  \\ 
		$112\times 112$&$8\times 8$&9 & 57.5 \\
		$112\times 112$&$8\times 8$&12 & 57.5\\
		$112\times 112$&$112\times 112$&200 & 55.8\\
		$112\times 112$&$112\times 112$&300 & 56.4\\
		\Xhline{2\arrayrulewidth}
	\end{tabular}
	\label{table: detail}
% \end{table}
\end{wraptable}

\begin{figure}[!tb]
\centering
    \subfloat[elements of fg patches]{
		\label{fig: fg}
		\includegraphics[width=0.32\textwidth,height=0.13\textheight]{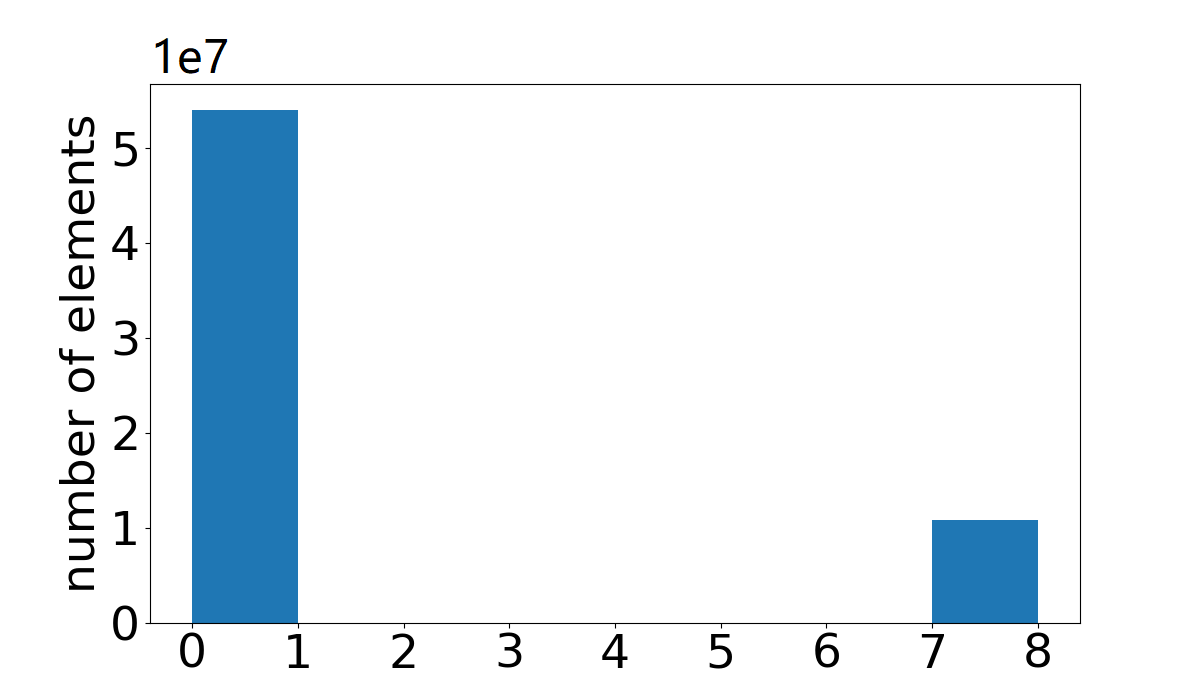}
	}
	\subfloat[elements of bg patches]{
		\label{fig: dcc}
		\includegraphics[width=0.32\textwidth,height=0.13\textheight]{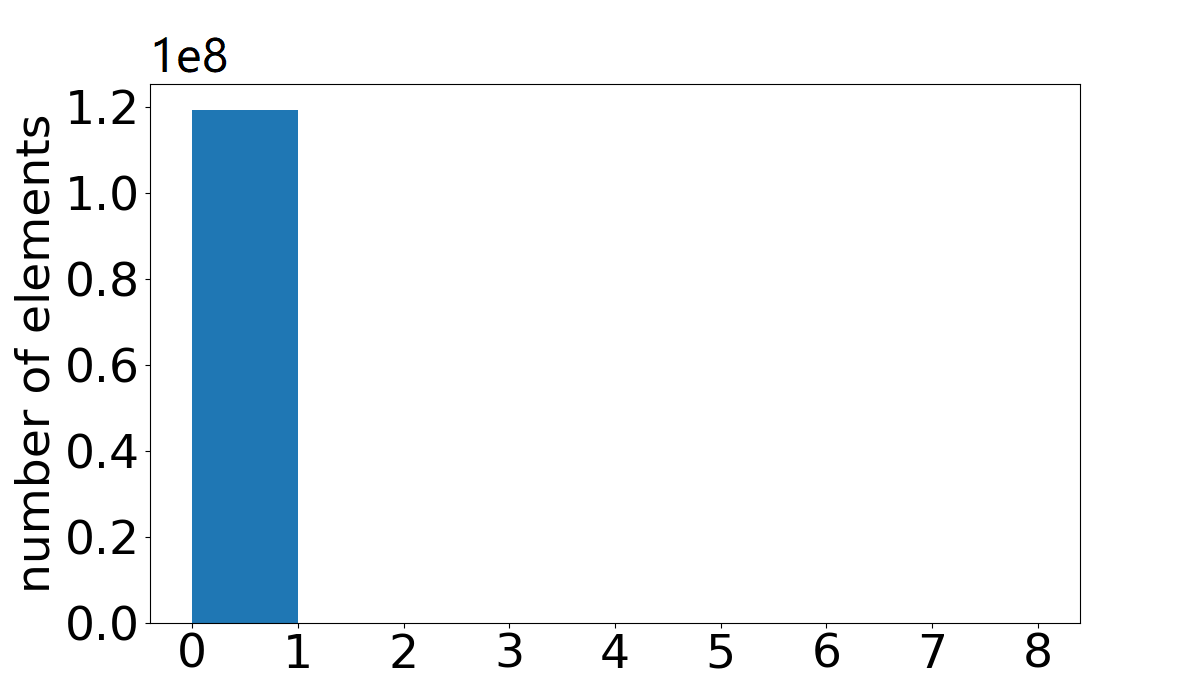}
	}
	\subfloat[elements of mixed patches]
	{
		\label{fig: mixed dcc}
		\includegraphics[width=0.32\textwidth,height=0.13\textheight]{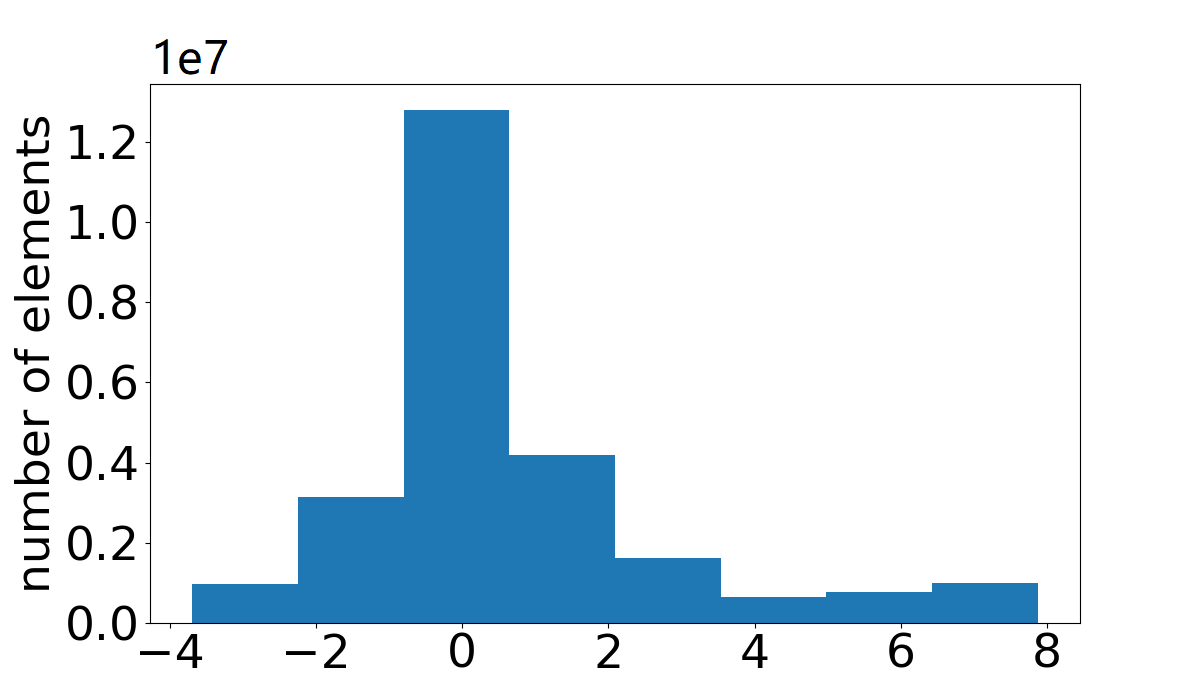}
	}
	%\subfloat[$2$-th element(all)]{
		%\label{fig: second}
		%\includegraphics[width=0.25\textwidth]{iclr2023/figure/c2a.png}
	%}
	%\subfloat[$2$-th elements(mixed)]
	%{
		%\label{fig: mixed second}
		%\includegraphics[width=0.25\textwidth]{iclr2023/figure/c2m.png}
	%}
	\vspace{2mm}
	\caption{Elements of $6$-dimensional DCT vectors for foreground, background and mixed patches on COCO \emph{val2017}. DCT vector elements for foreground and background patches are discrete data. DCT vector elements for mixed patches are continuous data.}
	\vspace{-2mm}
	\label{fig:dcthist}
\end{figure}

%\subsection{Necessity of the three-class classifier.}
\textbf{Three-class classifier.}
We define the patches with only foreground pixels and only background pixels as foreground patches and background patches, respectively, while the others are mixed patches. %Intuitively, the mixed patches are generally distributed on the boundary of object.
The task of differentiating patch categories is accomplished by a fully convolutional three-class classifier. 
%A fully convolutional three-class classier is used to differentiate patch categories. 
Moreover, the mispredicted initial foreground and background patches are corrected by the classifier.
%We utilizes a classifier to classify patches into 3 categories: foreground, background, and mixed patches. 
%We ultilizes a three-class classifier to distinguish each patch.
%The mispredicted initial foreground and background patches are corrected by the classifier, and the mixed patches will be further refined by the regressor.
We utilize a three-class classifier instead of a DCT vector regressor to refine foreground and background patches because of the particular form %special form 
of their DCT vectors. For background patches, simply from Equation \ref{eq: dct}, all elements of DCT vectors are zero. For foreground patches, all elements are zero except for the first element named DC component (DCC), which is equal to the patch size $m$. 
The mathematical proof of the DCT vector form for the foreground % and background
patches is shown in the Appendix.
DCT vector elements of foreground and background patches are discrete data that are more suitable for classification.
Referring to Figure \ref{fig:dcthist}, DCT vector elements of mixed patches are continuously distributed and therefore more suitable for regression.

\textbf{Regressor.}
Similar to the phenomenon described in DCT-Mask \citep{shen2021dct}, refining high-resolution masks with the binary grid mask representation introduces performance degradation due to the high training complexity (refer to DCT-Mask \citep{shen2021dct} for more details). Learning to regress informative DCT vectors eases the training process. The specific experimental results are discussed in the experiments section (Sec. \ref{sec:exp}).

The regressor is trained and inferred for mixed patches only. It is actually a boundary attention module, since the mixed patches are distributed exactly along the boundary of the instance mask. For each mixed patch, the regressor predicts an $n$-dimensional DCT vector, which is very short but highly informative.
%The $n$-dimensional DCT vectors can be very short ($n\ll N$) to reduce the training complexity, but still be highly informative.
Table \ref{table: detail} shows mask AP obtained by different lengths of ground truth patch DCT vectors using Mask-RCNN framework on COCO \emph{val2017}. 
The low-dimensional DCT vectors have been able to provide sufficient ground truth information.
%Low-dimensional DCT vectors can provide enough ground truth information. %Furthermore, the regressor is actually a boundary attention module, since the mixed patches are distributed exactly along the boundary of instance mask. 

%Even if an initial mask is provided, prediction of high resolution masks in binary grid representation still has a high training complexity and leads to bad prediction results \citep{shen2021dct}. 
%Similar to  the phenomenon described in DCT-Mask \citep{shen2021dct}, refining high resolution mask with binary grid representation introduces high training complexity and performance degradation. 
%The specific experimental results will be discussed in the Experiments
%section (Sec. \ref{sec:exp}).
%Regressing DCT vectors of mixed patches allows PatchDCT to easily generate fine boundaries with lower training complexity.

\subsection{Multi-stage Refinement and Loss Function}
PatchDCT is a module where the input and output masks have the same resolution. Thus, the mask generated by a PatchDCT module can be fed into another PatchDCT module for further refinement, as shown in the upper right corner of Figure \ref{fig:patchdctmodel}.

With multi-stage refinement, the loss function of the mask branch is defined as %follow:
\begin{align}
	\mathcal{L}_{mask} = \lambda_0\mathcal{L}_{dct_{N}}+\sum_{s> 0}\lambda_s(\mathcal{L}^s_{cls_{patch}}
	+\mathcal{L}^s_{dct_n}),
	\label{eq:loss}
\end{align}
$\lambda_0$ and $\lambda_s$ are the loss weights.
The first item $\mathcal{L}_{dct_{N}}$ of Equation \ref{eq:loss} is the loss in predicting $N$-dimensional vectors of the entire masks \citep{shen2021dct}.
\begin{align}
	\mathcal{L}_{dct_{N}} = \frac{1}{N}\sum_i^{N} R(\hat{V}_i-V_i), \space 
\end{align}
where $V_i$ and $\hat{V_i}$ are the $i$-th element in ground-truth and the prediction vector respectively.
$R$ is the loss function and $N$ is the length of the vectors. 
The classification loss $\mathcal{L}^s_{cls_{patch}}$ of $s$-th stage is the cross-entropy loss over three classes. The regression loss $\mathcal{L}^s_{dct_{n}}$ of $s$-th stage is
\begin{align}
	\mathcal{L}^s_{dct_{n}} = \frac{1}{N_{m}}\sum_{k}^{N_{all}}
	\left[p^k\left(\frac{1}{n}\sum_i^{n} R(\hat{V}_i-V_i) \space \right)\right], 
\end{align}
where $N_{m}$, $N_{all}$ are the number of mixed patches and all patches respectively.
$n$ is the length of the patch DCT vectors.
If the $k$-th patch is a mixed patch,  $p^k=1$, otherwise $p^k=0$, indicating that only DCT vectors of mixed patches are regressed.

\section{Experiments}\label{sec:exp}

\subsection{Datasets}
We evaluate our method on two standard instance segmentation datasets: COCO \citep{lin2014microsoft} and Cityscapes \citep{cordts2016Cityscapes}. 
COCO provides $80$ categories with instance-level annotations. 
%Our model is trained on COCO \emph{train2017} %(around 118k images) and validated on \emph{val2017} and \emph{test-dev2017}.
Cityscapes is a dataset focused on urban street scenes. 
It contains $8$ categories for instance segmentation, providing  
2,975, 500 and 1,525 high-resolution images ($1,024\times 2,048$) 
for training, validation, and test respectively.

We report the standard mask AP metric and the Boundary AP \citep{cheng2021boundary} metric (AP$_{B}$), the latter focusing on evaluating the boundary quality.
%which focuses on evaluating the boundary quality.
%Since COCO annotations are very coarse, LVIS\cite{gupta2019lvis} re-annotates COCO dataset and provides ground truth with higher quality annotations. 
Following \citep{kirillov2020pointrend}, we also report AP$^*$ and AP$^*_B$, which evaluate COCO \emph{val2017}  with high-quality annotations provided by LVIS \citep{gupta2019lvis}. Note that for AP$^*$ and AP$^*_B$, models are still trained on COCO \emph{train2017}.

\subsection{Implement Details}

%DCT-Mask\cite{shen2021dct} is the baseline for our model in which only its mask branch is modified. 
We build the model based on DCT-Mask \citep{shen2021dct}. 
%We first predict a $300$-dimensional vector and decode it to obtain a $112\times 112$ mask by Equation~\ref{eq: idct}.
We first decode the $300$-dimensional DCT vector to obtain a  $112 \times 112$ mask.
%A 112 $\times$ 112 mask is firstly obtained by decoding from a $300$-dimensional DCT vector.
%A $112\times 112$ mask is decoded from a $300$-dimensional vector
%and then downsample into $42\times 42$ through bilinear interpolation. 
This mask is then fed into PatchDCT, together with a $42\times 42$ feature map cropped from FPN-P2 \citep{lin2017feature}.
PatchDCT refines each patch of the mask and outputs a $112\times 112$ mask.
%The implementation of PatchDCT is shown in Figure~\ref{fig:model}.
%We set patch size $m$ to $8$. 
We set the patch size to $8$ and each patch is represented by a $6$-dimensional DCT vector.
Our model is class-specific by default, i.e. one mask per class.
$L1$ loss and cross-entropy loss are used for DCT vector regression and patch classification respectively.
%We use $L1$ loss in DCT vector regression and cross-entropy loss in patch classification. 
%since $6$-dimensional DCT vectors already provide enough information. Table~\ref{table: detail} shows mask AP obtained for different dimensions of the ground truth patch DCT vectors using Mask-RCNN framework on COCO \emph{val2017} dataset. More elements do not yield a significant increment of information, but rather increase the computational cost.
By default, only one PatchDCT module is used, and both $\lambda_0$ and $\lambda_1$ are set to 1.
%As we only uses one PatchDCT module, we set both $\lambda_0$ and $\lambda_1$ to 1.
%For multi-stage PatchDCT, PatchDCT module of each stage shares the same parameters.
%We use two-stage PatchDCT in the paper and set $\lambda_0$, $\lambda_1$, $\lambda_2$ to 0.5, 0.5, 0.8 respectively.x
We implement our algorithm based on Detectron2 \citep{wu2019detectron2}, and all hyperparameters remain the same as Mask-RCNN in Detectron2.
Unless otherwise stated, $1\times $ learning schedule is used.
%in Detectron2.
\subsection{Main Results}
%Our model is compared with Mask-RCNN and DCT-Mask with R50-FPN.
%We compare with Mask-RCNN and DCT-Mask RCNN, as well as other state-of-the-art methods.

\textbf{Results on COCO.}
We compare PatchDCT with Mask-RCNN and DCT-Mask over different backbones. As shown in Table \ref{table: backbones}, %Table \ref{table: backbones} lists the results obtained by different backbones.
on COCO \emph{val2017} with R50-FPN, PatchDCT improves 2.0\% AP and 3.4\% AP$_{B}$ over Mask-RCNN. Compared with DCT-Mask, PatchDCT also achieves 0.7\% AP and 0.9\% AP$_{B}$ improvements. % over DCT-Mask.
%With LVIS annotations, 
When evaluating with LVIS annotations, %PatchDCT obtains more improvements. 
PatchDCT yields significant gains of 3.2\% AP$^*$ and 5.3\% AP$^*_B$ over Mask-RCNN, and 1.1\% AP$^*$  and 1.7\% AP$^*_B$ over DCT-Mask.
Consistent improvements are observed on R101-FPN and RX101-FPN.
Since AP$^*$ and AP$^*_B$ are evaluated with high-quality annotations, the significant improvements of these two metrics emphasize the superiority of our model.
%Recalling that AP$^*$ and AP$^*_B$ are evaluated with high quality annotations, which better reflect the ability of predicting high-quality mask.
%PatchDCT surpasses Mask-RCNN by 3.2\% AP$^*$ and 5.3\% AP$^*_B$. % on LVIS. 
%It increases DCT Mask by 1.4\% AP$^*$  and 1.7\% AP$^*_B$.
In addition, considering the improvement in mask quality, the cost in runtime is almost negligible, i.e. about 1.5 FPS degradation on the A100 GPU.
%Besides, considering the improvements in mask performance, the cost in runtime is almost negligible, i.e. about 1.5 FPS degradation on the A100 GPU.
%Moreover, PatchDCT achieves consistent improvements on R101-FPN and RX101-FPN.  
%\wqr{PatchDCT is little slower than Mask-RCNN and DCT-Mask.%However, considering the improvement of masks, the cost in the runtime is almost negligible.}

%PatchDCT also achieves steady improvements over Mask-RCNN and DCT-Mask on both mAP and Boundary AP.

\begin{table*}[tb!]
\begin{center}
     \caption{Mask AP on COCO with different backbones based on Mask-RCNN framework. AP$^*$ is results obtained from COCO with LVIS annotations. AP$_{B}$ is Boundary AP. AP$_{B}^*$ is Boundary AP using LVIS annotations. Models with R101-FPN and RX101-FPN are trained with `3$\times$' schedule.
     Runtime is measured on a single A100.
     Considering the significant improvement of masks, the cost in the runtime is almost negligible.}
	\scalebox{0.82}{
		
		\begin{tabular}{l|l|c|ccc|c|c|ccc|c|c}
			\Xhline{2\arrayrulewidth}
			\textbf{Backbone}    & \textbf{Model} & \textbf{AP}&\textbf{AP$_S$}&
			\textbf{AP$_M$}&\textbf{AP$_L$}&\textbf{AP$_{B}$} & \textbf{AP$^*$}&\textbf{AP$_S^*$}&
			\textbf{AP$_M^*$}&\textbf{AP$_L^*$}&\textbf{AP$_{B}^*$}&\textbf{FPS}\\
			\hline
			\multirow{3}{*}{R50-FPN}   & Mask-RCNN &35.2&17.2&37.7&50.3 &21.1&37.6 & 21.3&	43.7 &	55.1 &24.8&\textbf{13.9}\\
			& DCT-Mask&36.5&17.7&38.6&51.9&23.6&39.7&\textbf{23.5}&46.5&58.5 &28.4&13.2\\ 
			&PatchDCT&\textbf{37.2}&\textbf{18.3}&\textbf{39.5}&\textbf{54.2}&\textbf{24.5}&\textbf{40.8}&23.0&\textbf{47.7}&\textbf{60.7}&\textbf{30.1}&12.3\\
			%&two-stage PatchDCT&\textbf{37.4}&\textbf{17.8}&\textbf{40.0}&\textbf{54.0}&\textbf{24.7}&\textbf{41.2}&22.6&\textbf{48.2}&\textbf{61.2}&\textbf{30.5}\\
			\hline
			\multirow{3}{*}{R101-FPN}   & Mask-RCNN &38.6 &19.5&41.3&55.3
			&24.5&41.4& 24.5 &47.9 & 61.0&29.0&\textbf{13.8}\\
			& DCT-Mask&39.9&20.2&42.6&57.3&26.8&43.7&25.8&50.5 &64.6&32.4&13.0\\ 
			& PatchDCT& \textbf{40.5}&\textbf{20.8} & \textbf{43.3}& \textbf{57.7}&\textbf{27.6}&\textbf{44.4}&\textbf{27.0}&\textbf{51.5}&\textbf{65.3}&\textbf{33.8}&11.8\\
			%& two-stage PatchDCT& \textbf{40.7}&\textbf{20.6} & \textbf{43.4}& \textbf{58.5}&\textbf{27.8}&\textbf{44.7}&\textbf{25.9}&\textbf{52.0}&\textbf{65.2}&\textbf{34.1}\\
			\hline
			\multirow{3}{*}{RX101-FPN}   & Mask-RCNN &39.5&20.7&42.0&56.5
			&25.3&42.1&25.4  & 48.0&61.4&29.7&\textbf{13.3}\\
			& DCT-Mask&41.2&21.9&44.2&57.7
			&28.0&45.2 &27.4&52.6&64.2&34.0&12.9\\ 
			& PatchDCT
			&\textbf{41.8}&\textbf{22.5}&\textbf{44.6}&\textbf{58.7}&\textbf{28.6}&\textbf{46.1}&\textbf{27.8} &\textbf{53.0} &\textbf{66.1}&\textbf{35.4}&11.7\\
			\Xhline{2\arrayrulewidth}
	\end{tabular}}
	\label{table: backbones}
\end{center}
\end{table*}

\begin{table}[tb!]
    \begin{center}
    \caption{Results on Cityscapes \emph{val} set.
	AP$_{B}$ is Boundary AP. 
	All models are based on R50-FPN backbone.
	PatchDCT achieves the best performance.}
	\scalebox{0.95}{
		\begin{tabular}{l|c|c|c|c}
		\Xhline{2\arrayrulewidth}
		%\multirow{2}{*}{Methods}&\multicolumn{5}{c|}{COCO}&Cityscapes\\
		\textbf{Methods}&\textbf{Resolution}&\textbf{AP}&\textbf{AP$_{50}$}&\textbf{AP$_{B}$} \\ 
		\hline
			Mask-RCNN \citep{he2017mask}&$28\times 28$&33.7&60.9&11.8\\
			\wqr{Panoptic-DeepLab \citep{cheng2020panoptic}} & -& 35.3& 57.9&16.5\\
			PointRender \citep{kirillov2020pointrend}&$224\times 224$&35.9&61.8&16.7\\
			DCT-Mask \citep{shen2021dct}&$112\times 112$&36.9&62.9&14.6\\
			RefineMask \citep{zhang2021refinemask}&$112\times112$&37.6&63.3&17.4\\
			%Mask Transfiner\cite{ke2022mask}&37.3&24.2&40.5&-&-& 37.9\\
			Mask Transfiner \citep{ke2022mask}&$112\times112$&37.9&64.1&18.0\\
			\hline
			PatchDCT&$112\times 112$&\textbf{38.2}&\textbf{64.5}&\textbf{18.8}\\
			%Two-stage PatchDCT&$112\times 112$&37.8&63.3&\textbf{18.8}\\
			\Xhline{2\arrayrulewidth}
	\end{tabular}}
	\label{table: cityscapes}
\end{center}
% \vspace{-3mm}
\end{table}

We also compare the performance of PatchDCT with state-of-the-art methods of instance segmentation on COCO \emph{test-dev2017}. 
%All comparison results are listed in Table \ref{table:sota}.
%PatchDCT obtains better performance in comparison with PointRend \citep{kirillov2020pointrend}, which first obtains $7\times 7$ mask and renders it to $224\times224$. 
%PatchDCT achieves comparable performance with Mask Transfiner \citep{ke2022mask} without using the computationally more expensive Transformer. 
With RX101 backbone, PatchDCT surpasses  PointRender \citep{kirillov2020pointrend} and RefineMask \citep{zhang2021refinemask}, which are both multi-stage refinement methods based on binary grid masks, by 0.8\% and 0.4\%.
PatchDCT also achieves comparable performance with Mask Transfiner \citep{ke2022mask} with R101 backbone. However, Mask-Transifer runs at 5.5 FPS on the A100 GPU, which is almost two times slower than PatchDCT.
%\wqr{PatchDCT achieves comparable performance with Mask Transfiner \citep{ke2022mask}, while the former method is nearly two times faster than the latter.
%The fps of Mask Transfiner on R101 is 5.5, measured on the same A100 in Table \ref{table: backbones}.}
%With RX101 backbone, PatchDCT surpasses RefineMask \citep{zhang2021refinemask}, which utilizes a multi-stage procedure to generates $112\times 112$ binary grid masks. 
%The results demonstrate that PatchDCT is an effective refinement module for high-resolution masks.
\wqr{
With Swin-B backbone, PatchDCT outperforms Mask Transfiner \citep{ke2022mask} by 0.7\% AP.
It is worth noting that PatchDCT is faster than most multi-stage refinement methods since only one refine process is required.}
%With the help of highly informative patch DCT vectors, PatchDCT significantly improves the segmentation quality with only one refinement of the mask.}
These results demonstrate the effectiveness of PatchDCT in generating high-quality masks.

\textbf{Results on Cityscapes.} 
We also report results on Cityscapes \emph{val} set in Table \ref{table: cityscapes}. 
In comparison with Mask-RCNN, PatchDCT obtains 4.5\% AP and 7.0\% AP$_{B}$ improvements.
It also outperforms DCT-Mask by 1.3\% AP and 4.2\% AP$_{B}$.
%PatchDCT also gains improvements on both AP and AP$_{B}$ compared with other SOTA methods.
Compared with other SOTA methods, PatchDCT is still competitive. %PatchDCT yields high-quality boundaries.
PatchDCT achieves 0.8\%, 1.4\%, 2.1\% AP$_{B}$ gains over Mask Transfiner \citep{ke2022mask}, RefineMask \citep{zhang2021refinemask} and PointRender \citep{kirillov2020pointrend} respectively. 
The large difference in AP$_{B}$ highlights the ability of PatchDCT to generate masks with more detailed borders.

\subsection{Ablation Experiments}
\begin{table*}[tb!]
	\centering
 \caption{Comparison of different methods on COCO \emph{test-dev2017}. MS denotes using multi-scale training. `3$\times$' schedules indicates 36 epochs for training. \wqr{Runtime is measured on a single A100.}}
	\scalebox{0.75}{
		\begin{tabular}{l|l|c|c|c|cc|ccc|r}
			\Xhline{2\arrayrulewidth}
			\textbf{Method}&\textbf{Backbone}&\textbf{MS}&\textbf{Sched.}&\textbf{AP}&\textbf{AP$_{50}$}&\textbf{AP$_{75}$}&\textbf{AP$_S$}&
			\textbf{AP$_M$}&\textbf{AP$_L$}&\wqr{FPS}
			\\
			\hline
			BMask RCNN \citep{cheng2020boundary}&R101-FPN& &1$\times$&37.7&59.3&40.6& 16.8&39.9&54.6&-\\
			Mask-RCNN \citep{he2017mask}&R101-FPN&\checkmark&3$\times$&38.8&60.9&41.9&21.8&41.4& 50.5&\wqr{13.8}\\
			
			BCNet \citep{ke2021deep}&R101-FPN&\checkmark&3$\times$&39.8&61.5&43.1 &22.7&42.4&51.1&\wqr{-}\\
			DCT-Mask \citep{shen2021dct}&R101-FPN&\checkmark&3$\times$&40.1&61.2&43.6&22.7&42.7&51.8&\wqr{13.0}\\
			Mask Transfiner \citep{ke2022mask}&R101-FPN&\checkmark&3$\times$&40.7&-&-&23.1&42.8&53.8&\wqr{5.5}\\
			SOLQ \citep{dong2021solq}&R101-FPN&\checkmark&50e&40.9&-&-&22.5&43.8&54.6&\wqr{10.7}\\
			MEInst \citep{zhang2020mask}&RX101-FPN&\checkmark&3$\times$&36.4&60.0&38.3&21.3& 38.8&45.7&\wqr{-}\\
			HTC \citep{chen2019hybrid}&RX101-FPN& &20e&41.2&63.9&44.7&22.8&43.9&54.6&\wqr{4.3}\\
			PointRend \citep{kirillov2020pointrend}&RX101-FPN&\checkmark&3$\times$&41.4&63.3&44.8&24.2&43.9&53.2&\wqr{8.4}\\
			RefineMask \citep{zhang2021refinemask}&RX101-FPN&\checkmark&$3\times$&41.8&-&-&-&-&-&\wqr{8.9}\\
			\wqr{Mask Transfiner \citep{ke2022mask}}&Swin-B&\checkmark&3$\times$&45.9&69.3&50.0&28.7&48.3&59.4&3.5 \\
			
			\hline
			PatchDCT&R101-FPN&\checkmark&3$\times$&40.7&61.8&44.2&22.8&43.2&52.8&\wqr{11.8}\\
			PatchDCT&RX101-FPN&\checkmark&3$\times$&42.2&64.0&45.8&25.0&44.5&53.9&\wqr{11.7} \\
			\wqr{PatchDCT}&Swin-B&\checkmark&3$\times$&\textbf{46.6}&\textbf{69.7}&\textbf{50.8}&\textbf{29.0}&\textbf{49.0}&\textbf{59.9}&7.3
			
			\\
			%\hline
			
			\Xhline{2\arrayrulewidth}
			
	\end{tabular}}
	\label{table:sota}
 % \vspace{-3mm}
\end{table*}

\begin{table}[t]
\begin{minipage}[t]{0.48\linewidth}
\caption{Mask AP obtained by different refinement methods on \emph{val2017}. PatchDCT significantly improves the quality of masks.}
\centering
	\scalebox{0.7}{
		\begin{tabular}{l|c|c|c|c}
			\Xhline{2\arrayrulewidth}
			\textbf{Method}&\textbf{AP}&\textbf{AP$_{B}$}&\textbf{AP$^*$}&\textbf{AP$_{B}^*$}
			\\
			\hline
			Binary grid&35.7&23.2&39.6&29.1
			\\
			Two-stage DCT&36.6&23.9&40.1&29.1
			\\
			PatchDCT&\textbf{37.2}&\textbf{24.7}&\textbf{40.8}&\textbf{30.1}
			\\ 
			\Xhline{2\arrayrulewidth}
	\end{tabular}}
	\label{table:multi-stage dct}
\end{minipage}
\quad
\begin{minipage}[t]{0.48\linewidth}
\caption{Mask AP obtained by PatchDCT with two-class classifier and three-class classifier on \emph{val2017}. PatchDCT with three-class classifier achieves the best performance.}
\centering
\scalebox{0.7}{
\begin{tabular}{l|c|ccc|c|c|c}
\Xhline{2\arrayrulewidth}
\textbf{Classifier}  & \textbf{AP} &\textbf{AP$_S$}&
\textbf{AP$_M$}&\textbf{AP$_L$}&\textbf{AP$_{B}$}&\textbf{AP$^*$}&\textbf{AP$_{B}^*$}\\
\hline
2-class&36.9&18.2&39.3&53.5&24.4&40.4&29.7\\
3-class &\textbf{37.2}&\textbf{18.3}&
\textbf{39.5}&\textbf{54.2}&\textbf{24.5}&\textbf{40.8}&\textbf{30.1}\\ 
\Xhline{2\arrayrulewidth}
\end{tabular}}
\label{table: classifier}
\end{minipage}

\vspace{3mm}

\begin{minipage}[t]{0.48\linewidth}
\caption{Mask AP obtained by PatchDCT with regressor focusing on all patches and mixed patches on \emph{val2017}. The best results are obtained by regressing only the mixed patches.}
	\centering
		\scalebox{0.68}{
			\begin{tabular}{l|c|ccc|c|c|c}
				\Xhline{2\arrayrulewidth}
				\textbf{Regressor} & \textbf{AP} &\textbf{AP$_S$}&
\textbf{AP$_M$}&\textbf{AP$_L$}&\textbf{AP$_{B}$}&\textbf{AP$^*$}&\textbf{AP$_{B}^*$}\\
				\hline
				all &36.6&17.7&39.5&52.2&23.6&39.6&28.6\\
			    mixed&\textbf{37.2}&\textbf{18.3}&
				\textbf{39.5}&\textbf{54.2}&\textbf{24.5}&\textbf{40.8}&\textbf{30.1}\\ 
				
				\Xhline{2\arrayrulewidth}
		\end{tabular}}
		\label{table: mixed patches}
\end{minipage}
\quad
\begin{minipage}[t]{0.48\linewidth}
\caption{Mask AP obtained by PatchDCT with and without the regressor on \emph{val2017}. PatchDCT benefits from the regressor.}
\centering
		\scalebox{0.68}{
			\begin{tabular}{c|c|ccc|c|c|c}
				\Xhline{2\arrayrulewidth}
				\textbf{Regressor} & \textbf{AP} &\textbf{AP$_S$}&
\textbf{AP$_M$}&\textbf{AP$_L$}&\textbf{AP$_{B}$}&\textbf{AP$^*$}&\textbf{AP$_{B}^*$}
				\\
				\hline
				&36.7&18.3&39.0&53.1&23.3&39.6&27.1\\
				\checkmark&\textbf{37.2}&\textbf{18.3}&
				\textbf{39.5}&\textbf{54.2} &\textbf{24.5}&\textbf{40.8}&\textbf{30.1}\\ 
				
				\Xhline{2\arrayrulewidth}
		\end{tabular}}
		\label{table: regressor}
\end{minipage}

\vspace{3mm}

\begin{minipage}[t]{0.48\linewidth}
\caption{Mask AP obtained by models with different dimensions of patch DCT vectors on COCO \emph{val2017}. Model with $6$-dimensional vectors achieves the best performance. }
\centering
		\scalebox{0.68}{
			\begin{tabular}{c|c|ccc|c|c|c}
				\Xhline{2\arrayrulewidth}
				\textbf{Patch Dim.} & \textbf{AP} &\textbf{AP$_S$}&
\textbf{AP$_M$}&\textbf{AP$_L$}&\textbf{AP$_{B}$}&\textbf{AP$^*$}&\textbf{AP$_{B}^*$}
				\\
				\hline
				3&36.8&17.6&39.2&53.5&24.0&40.5&29.5\\	
				6&\textbf{37.2}&\textbf{18.3}&\textbf{39.5}&\textbf{54.1}&\textbf{24.5}&\textbf{40.8}&\textbf{30.1}\\ 
				9&36.9&17.1&39.3&53.3&24.3&40.6&30.1\\  
				%12&&&& \\
				\Xhline{2\arrayrulewidth}
		\end{tabular}}
		\label{table: dimension}
\end{minipage}
\quad
\begin{minipage}[t]{0.48\linewidth}
\caption{Mask AP obtained by multi-stage PatchDCT on \emph{val2017}. Two-stage PatchDCT achieves a trade-off between accuracy and computational complexity.}
\centering
		\scalebox{0.62}{
			\begin{tabular}{c|c|ccc|c|c|c|c}
				\Xhline{2\arrayrulewidth}
				\textbf{Stage}  & \textbf{AP} &
				\textbf{AP$_S$}&\textbf{AP$_M$}&\textbf{AP$_L$}&
				\textbf{AP$_{B}$}&\textbf{AP$^*$}
				%&AP$_{B}^*$
				&\textbf{(G)FLOPs}&\wqr{FPS}
				\\
				\hline
				1 &37.2
				&\textbf{18.3}&39.5&54.1
				&24.5&40.8
				%&30.1
				&\textbf{5.1}&\wqr{12.3}\\ 
				2 &\textbf{37.4}
				&17.8&\textbf{40.0}&54.0
				&\textbf{24.7}&\textbf{41.2}
				%&\textbf{30.5}
				&9.6&\wqr{11.1}\\
				3&37.3
				&17.3&39.7&\textbf{54.6}
				&\textbf{24.7}&40.9
				%&30.4
				&14.1&\wqr{8.4}\\
				
				\Xhline{2\arrayrulewidth}
		\end{tabular}}
		\label{table: multi-stage}
            \vspace{3mm}

\end{minipage}

\begin{minipage}[t]{0.48\linewidth}
\caption{Mask AP obtained by models with different patch sizes on COCO \emph{val2017}. PatchDCT with $8\times 8$ patch size obtains the best performance.}
\centering
		\scalebox{0.68}{
			\begin{tabular}{c|c|ccc|c|c|c}
				\Xhline{2\arrayrulewidth}
				\textbf{Patch Size}& \textbf{AP} &\textbf{AP$_S$}&
\textbf{AP$_M$}&\textbf{AP$_L$}&\textbf{AP$_{B}$}&\textbf{AP$^*$}&\textbf{AP$_{B}^*$}
				\\
				\hline
				$4\times 4$&37.0&17.5&39.3&53.8&24.4&40.5&29.8\\
				$8\times 8$&\textbf{37.2}&\textbf{18.3}&\textbf{39.5}&\textbf{54.1}&\textbf{24.5}&\textbf{40.8}&\textbf{30.1}\\ 
				$16\times 16$&37.0 &17.6 &39.3 &53.5&24.4&\textbf{40.8}&30.0\\  
				\Xhline{2\arrayrulewidth}
		\end{tabular}}
		\label{table: patchsize}
\end{minipage}
\quad
\begin{minipage}[t]{0.48\linewidth}
\caption{Mask AP obtained by models with different feature map sizes on COCO \emph{val2017}. The performance saturates with the $42\times 42$ feature map.}
\centering
		\scalebox{0.68}{
			\begin{tabular}{c|c|ccc|c|c|c}
				\Xhline{2\arrayrulewidth}
				\textbf{Feature Size} & \textbf{AP} &\textbf{AP$_S$}&
\textbf{AP$_M$}&\textbf{AP$_L$}&\textbf{AP$_{B}$}&\textbf{AP$^*$}&\textbf{AP$_{B}^*$}
				\\
				\hline
				$28\times 28$&37.1&17.8 &39.3 &53.4&24.5&40.6&30.0 \\	
				$42\times 42$&\textbf{37.2}&\textbf{18.3}&\textbf{39.5}&\textbf{54.1}&\textbf{24.5}&40.8&\textbf{30.1}\\ 
				$56\times 56$&37.0 &17.4 & 39.2&53.0&24.4&\textbf{41.0} &30.3\\  
				\Xhline{2\arrayrulewidth}
		\end{tabular}}		
		\label{table: featuremapsize}
\end{minipage}
\vspace{-2mm}
\end{table}

\begin{table}[tb!]
    \begin{center}
    \caption{Mask AP obtained by PatchDCT with the feature map cropped from all levels and P2 only on COCO \emph{val2017}. Model with the feature map of P2 obtains higher mAP.}
		\scalebox{0.85}{
			\begin{tabular}{c|c|ccc|c|c|c}
				\Xhline{2\arrayrulewidth}
				\textbf{Feature} & \textbf{AP} &\textbf{AP$_S$}&
\textbf{AP$_M$}&\textbf{AP$_L$}&\textbf{AP$_{B}$}&\textbf{AP$^*$}&\textbf{AP$_{B}^*$}\\
				\hline
				P2&\textbf{37.2}&\textbf{18.3}&\textbf{39.5}&\textbf{54.1}&\textbf{24.5}&\textbf{40.8}&\textbf{30.1}\\ 
				P2-P5&37.1&18.2&39.3&53.3&24.4&40.6&29.8 \\  
				\Xhline{2\arrayrulewidth}
		\end{tabular}}
		\label{table: FPN}
\end{center}
\vspace{-6mm}
\end{table}

	We conduct extensive ablation experiments to further analyze PatchDCT. 
	We adopt R50-FPN as the backbone %of Mask-RCNN framework 
	and evaluate the performance on COCO \emph{val2017}.
	%By default, we use one-stage PatchDCT(short for PatchDCT) in ablation experiments.
	
	%We try different approaches to refine the high-resolution masks and analyze the effectiveness of PatchDCT. We list the result in Table~\ref{table:multi-stage dct}
	
	\textbf{Simply refine DCT vectors.}
	Simply refining the global DCT vectors does not succeed. To demonstrate that, we design a model named  `Two-stage DCT', which regresses a new $300$-dimensional DCT vector after fusing the initial mask with a $42\times 42$ feature map from FPN-P2. The refined mask is decoded from the final DCT vector.  
	%In the inference time, we decode the second DCT vector and obtain the entire mask directly. 
	From Table \ref{table:multi-stage dct}, Two-stage DCT achieves only little improvements over DCT-Mask, since changes in some elements of the global DCT vector may affect the entire mask, even for the correct segmentation areas. PatchDCT leverages the patching mechanism to overcome this issue and outperforms Two-stage DCT by 1.0 AP$_{B}^*$.
	%which proves it's very difficult to obtain finer masks by regressing a finer $300$-dimensional DCT vector.
	%As mentioned in Sec. \ref{sec:diff}, such combination fails to allow the model to focus on areas that are previously mispredicted, and correctly predicted areas may be influenced by the refinement. %Therefore, it's important to refine each patch respectively.
	% As it's important to reduce the interaction of regional refinement, PatchDCT refines each patch respectively.
	
	\textbf{Binary grid refinement.}
	Refining masks with the binary grid mask representation can be considered as the extreme patching mechanism, which treats each pixel as a patch. However, simply refining high-resolution masks with the binary grid mask representation introduces performance degradation. We construct an experiment named `binary grid refinement', which predicts another $112\times 112$ mask with the binary grid mask representation after fusing the initial mask as well as a $56\times 56$ feature map from FPN-P2. Experimental results in Table \ref{table:multi-stage dct} show that the performance of binary grid refinement is worse than PatchDCT, and even DCT-Mask. This is because binary grid refinement requires the refinement module to learn 12544 ($112\times 112$) outputs, while PatchDCT only needs to learn at most 1176 ($14\times 14 \times 6$) outputs, which reduces the training complexity. 
	%We analyze whether it's feasible to refine a high-resoluton mask with binary grid representation. %when the initial mask is provided.
	%We try to generate another $112\times 112$ mask with binary grid representation after fusing the information of the initial mask, which we name `binary grid refinement'.
	%We fuse the mask with a $56\times 56$ feature map from FPN-P2,with the same operation as PatchDCT, followed by a head the same as Mask-RCNN.
	%However, even if an initial mask is provided, it’s still very difficult to predict the binary grid representation of a high-resolution mask because of the high training complexity \citep{shen2021dct}.
	%The quality of masks is even worse than the initial masks.
	%Thus, PatchDCT still adopt DCT on mixed patches to generate high-resolution masks.

	%\textbf{Regressing the mixed patches only}
	%We show the importance of having the regressor regress only DCT vectors of mixed patches.The results in Table~\ref{table: mixed patches} demonstrates that predicting DCT vectors of mixed patches generates finer masks. As the DCT vectors of foreground and background have special forms, regression that considers DCT vectors of all patches changes the distributions of regression targets and leads to little improvements over initial masks.

	\textbf{Effectiveness of three-class classifier.}
	%We further analyze the role of  the classifier.
	%The three-class classifier is designed to identify mixed patches as well as refine foreground and background patches.
	%For identifying mixed patches, it seems to be a choice to discard the classifier and adopt rule-based method. Specially, we evaluate each patch of the initial mask and determine the patch that contains both foreground and background pixels as mixed-patch. However, as shown in Table \ref{table: classifier}, the rule-based method is inferior to our designed three-class classifier by 0.4 AP and 0.7 AP$^*$, since the initial mask is so imprecise that cannot accurately distinguish boundaries. In addition, binary-class classifier, which only classify patches as mixed/non-mixed and keeps non-mixed patch unchanged, seems to be an alternative option. It is indeed superior to rule-based approach. However, compared to three-class classifier, it only distinguishes the mixed-patches and ignores the refinement of previously mis-predicted foregroud patch and background patch. Therefore, binary-class classifier is 0.3 AP lower than our three-class classifier method. 
	In addition to identifying mixed patches, a more important role of the three-class classifier is to correct previously mispredicted foreground and background patches. To validate the effectiveness of refining non-mixed patches (i.e. foreground and background patches), we construct a binary-class classifier, which only classifies patches as mixed or non-mixed and keeps masks of non-mixed patches unchanged. As shown in Table \ref{table: classifier}, the binary-class classifier is inferior to our three-class classifier by 0.3\% AP and 0.4\% AP$^*$, since the refinement of previously incorrectly predicted foreground and background patches is ignored.
	
	Refinement of foreground and background patches can also be accomplished with the DCT vector regressor. However, as discussed in Sec. \ref{sec:patchdct}, the DCT vector elements of the non-mixed patches only involve zero and $m$, making it ineffective to learn the DCT vectors of all patches directly. As shown in Table \ref{table: mixed patches}, the performance of the method refining non-mixed regions with the DCT vector regressor is lower than the method using a three-class classifier by 0.6\% AP and 1.2\% AP$^*$. Need to note that, AP$_{B}$ and
	AP$^*_{B}$ decrease by 0.9\% and 1.5\% respectively, reflecting that learning to regress non-mixed patches also affects the prediction of boundaries. 

	\textbf{Effectiveness of the regressor.}
	The regressor is actually a boundary attention module that generates finer boundaries. As shown in Table \ref{table: regressor}, after removing the regressor and keeping only the classifier, the overall AP only decreases by 0.5\% , but AP$_{B}$ and
	AP$^*_{B}$ decrease by 1.2\%  and 3.0\%  respectively. The phenomenon demonstrates the importance of the regressor for generating finer boundaries.
	%The regressor generates finer boundaries of masks that greatly improve the performance of PatchDCT.
	%We demonstrate the necessity of the regressor by removing the regressor and only refining foregound and background patches by the classifier. 
	%The results in Table \ref{table: regressor} show that both AP and Boundary AP fall sharply without the regressor. %Although only a short $n$-dimensional vector needs to be regressed for each mixed patch, a high-resolution boundary is obtained by decoding each DCT vector during inference time.
	%As the resolution of mask is almost equivalent to the resolution of boundary,
	%PatchDCT obtains high-resolution masks by generating high-resolution boundaries.
	%Thus the regressor is essential for PatchDCT.
	%The regressor contributes a lot to the improvements of masks, especially boundaries of masks. 
	
	\textbf{Dimension of PatchDCT vectors}
	We look for an appropriate patch DCT vector length to encode each mixed patch.
	%We analyze PatchDCT with different dimensions of DCT vectors. 
	Results in Table \ref{table: dimension} show that the model with $6$-dimensional patch DCT vectors obtains the best performance. As also shown in Table \ref{table: detail}, the $6$-dimensional patch DCT vector already contains most of the ground truth information.
	As more elements bring only very little incremental information, regressing these elements does not improve the prediction.

	%\textbf{Size of input masks.}
	%We analyze the size of mask input in PatchDCT.The results are in Table~\ref{table: compressedmask}. The model performance benefits from input masks of the same size as output masks.
	
	%\textbf{Shared parameters of PatchDCT module.}
	%We compare the performance of multi-stage PatchDCT with shared and unshared parameters(Table ~\ref{table:shared}).Parameters are not shared by each PatchDCT module for the latter method. For two-stage PatchDCT, the model with shared parameters outperforms the model with unshared parameters.
	
	\textbf{Multi-stage PatchDCT.}
	We compare the performance of the multi-stage procedure in Table \ref{table: multi-stage}. 
	One-stage PatchDCT already provides high-quality masks, while two-stage PatchDCT further improves the prediction.
	However, the computational cost of the mask branch has nearly doubled with tiny improvements in the quality of masks, so we choose to use one-stage PatchDCT in our paper.
	%Besides, the performance is almost saturates with one-stage PatchDCT.
	%The performance saturates with two PatchDCT modules.
	%Between models using different numbers of PatchDCT modules, the cost of speed is almost negligible.

	\textbf{Size of the patch.}
	We evaluate the influence of patch size in Table \ref{table: patchsize}. 
	We keep the resolution of the mask and the size of the input feature map unchanged and compare the model performance 
	with different patch sizes. PatchDCT with $8\times 8$ patches performs better than other settings.
	
	\textbf{Size of the feature map.}
	We compare the model with different sizes of the feature map used in PatchDCT. 
	Table \ref{table: featuremapsize} illustrates that the performance saturates with the $42\times 42$ feature map.

	\textbf{Feature map from FPN.}
	We evaluate PatchDCT with the feature map cropped from all pyramid levels or P2.
	Table \ref{table: FPN} shows that PatchDCT benefits from the finer feature map of P2.

\begin{figure*}
	\centering
	%\subfloat[Ground Truth]{
		%\includegraphics[width=0.2\textheight]{figure/gt.png}%}
	\subfloat[Mask-RCNN]{
		\includegraphics[width=0.95\textwidth,height=0.12\textheight]{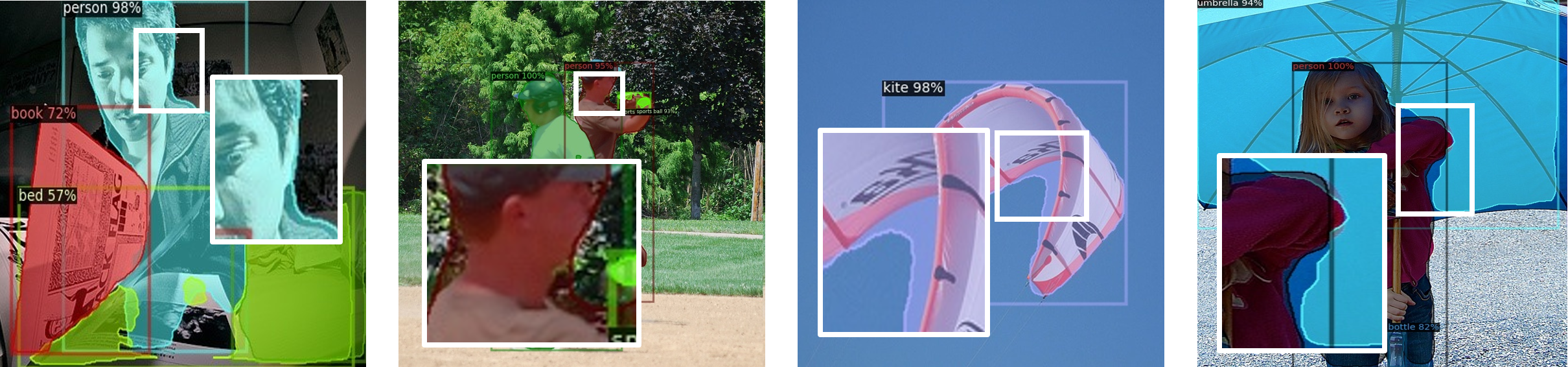}
	}
	\\
	\subfloat[DCT-Mask]{
		\includegraphics[width=0.95\textwidth,height=0.12\textheight]{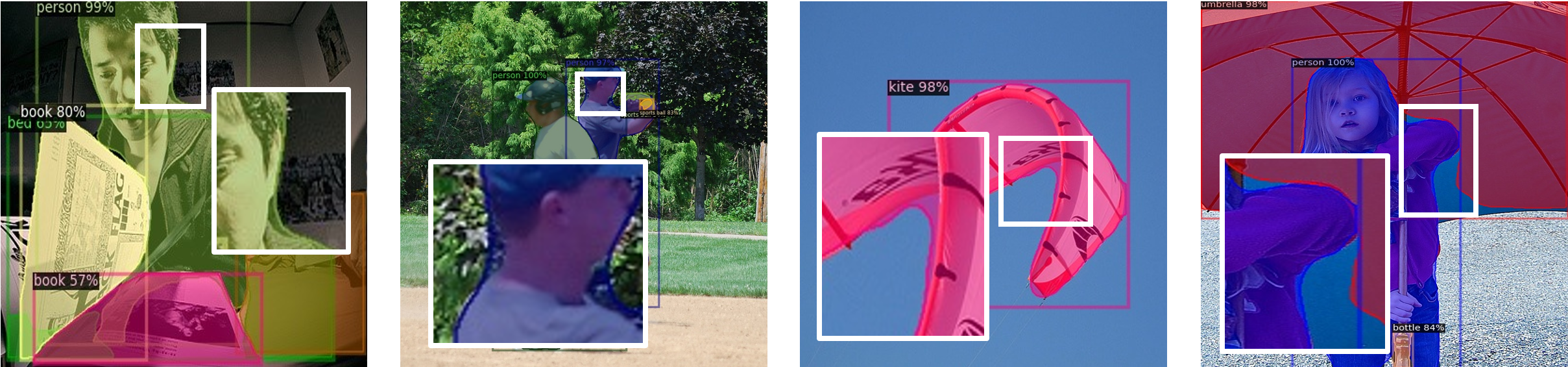}
	}
	%\subfloat[PatchDCT]{
		%\includegraphics[width=0.25\textwidth,height=0.45\textheight]{1p.png}
	%}
	\\
	\subfloat[PatchDCT]{
		\includegraphics[width=0.95\textwidth,height=0.12\textheight]{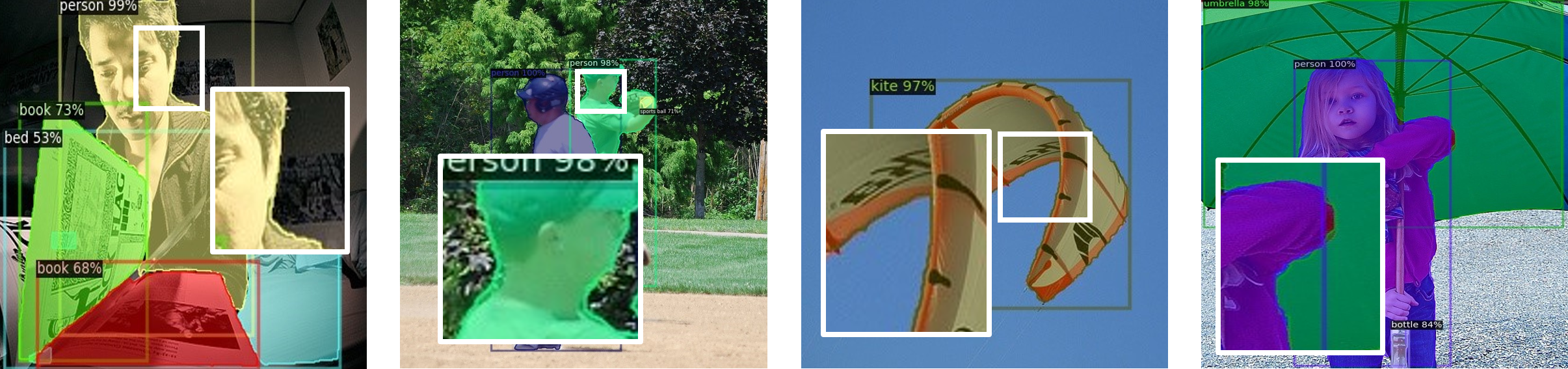}
	}
	\caption{COCO example tuples from 
		%the ground truth masks(the first column), 
		Mask-RCNN, DCT-Mask, and PatchDCT. Mask-RCNN, DCT-Mask and PatchDCT are trained based on R50-FPN. PatchDCT provides masks with higher quality and finer boundaries.}
	\label{fig:image}
 \vspace{-2mm}
\end{figure*}

	\subsection{Qualitative Results}
	
	In Figure \ref{fig:image} we visualize some outputs of PatchDCT on COCO \emph{val2017}. 
	PatchDCT generates finer boundaries among different instances, such as the shoulder of the person (the first column), the contour of the kite (the third column), and the arm of the girl (the fourth column).
	PatchDCT obtains masks of higher quality in comparison with Mask-RCNN and DCT-Mask.
	
	%In comparison with Mask-RCNN and DCT-Patch,PatchDCT provides masks with higher quality and finer boundary,such as …………
	
	\section{Conclusions}
	%In this work, we analyze the difficulties in refining DCT vectors and propose PatchDCT, a effective compressed vector based refinement module that generates high-quality masks. %with finer boundaries.
    %PatchDCT contains a three-class classifier and a DCT vector regressor.
    %The three-class refines foregound and background patches, and helps the regressor to focus on boundaries of masks.
    %By predicting DCT vectors of mixed patches, the regressor genenrates high-resolution boundaries.
    %PatchDCT can be further used iteratively to further improve the quality of masks.
	%We also discuss the relationship between the patch size and regression complexity of DCT vectors. Through mathematical formula, We demonstrate the characteristics of DCT vectors of pure foreground and background.
	In this work, we propose PatchDCT, a compressed vector based method towards high-quality instance segmentation.
    In contrast to previous methods, PatchDCT refines each patch of masks respectively and utilizes patch DCT vectors to compress boundaries that are full of details.
    By using a classifier to refine foreground and background patches, and predicting an informative low-dimensional DCT vector for each mixed patch, PatchDCT generates a high-resolution mask with fine boundaries.
    PatchDCT is designed with a simple and clean structure, which allows the method to obtain high-quality segmentation with almost negligible cost in speed compared to Mask-RCNN and DCT-Mask.
	We hope that our approach will benefit future studies in instance segmentation.
	%\CB{\{Conclusion is too simple!\}}

\bibliography{iclr2023_conference}

\begin{thebibliography}{40}
\providecommand{\natexlab}[1]{#1}
\providecommand{\url}[1]{\texttt{#1}}
\expandafter\ifx\csname urlstyle\endcsname\relax
  \providecommand{\doi}[1]{doi: #1}\else
  \providecommand{\doi}{doi: \begingroup \urlstyle{rm}\Url}\fi

\bibitem[Ahmed et~al.(1974)Ahmed, Natarajan, and Rao]{ahmed1974discrete}
Nasir Ahmed, T\_ Natarajan, and Kamisetty~R Rao.
\newblock Discrete cosine transform.
\newblock \emph{IEEE transactions on Computers}, 100\penalty0 (1):\penalty0
  90--93, 1974.

\bibitem[Al-Ani \& Awad(2013)Al-Ani and Awad]{al2013jpeg}
Muzhir~Shaban Al-Ani and Fouad~Hammadi Awad.
\newblock The jpeg image compression algorithm.
\newblock \emph{International Journal of Advances in Engineering \&
  Technology}, 6\penalty0 (3):\penalty0 1055, 2013.

\bibitem[Chen et~al.(2019)Chen, Pang, Wang, Xiong, Li, Sun, Feng, Liu, Shi,
  Ouyang, et~al.]{chen2019hybrid}
Kai Chen, Jiangmiao Pang, Jiaqi Wang, Yu~Xiong, Xiaoxiao Li, Shuyang Sun,
  Wansen Feng, Ziwei Liu, Jianping Shi, Wanli Ouyang, et~al.
\newblock Hybrid task cascade for instance segmentation.
\newblock In \emph{Proceedings of the IEEE/CVF Conference on Computer Vision
  and Pattern Recognition}, pp.\  4974--4983, 2019.

\bibitem[Chen et~al.(2017)Chen, Papandreou, Kokkinos, Murphy, and
  Yuille]{chen2017deeplab}
Liang-Chieh Chen, George Papandreou, Iasonas Kokkinos, Kevin Murphy, and Alan~L
  Yuille.
\newblock Deeplab: Semantic image segmentation with deep convolutional nets,
  atrous convolution, and fully connected crfs.
\newblock \emph{IEEE transactions on pattern analysis and machine
  intelligence}, 40\penalty0 (4):\penalty0 834--848, 2017.

\bibitem[Cheng et~al.(2020{\natexlab{a}})Cheng, Collins, Zhu, Liu, Huang, Adam,
  and Chen]{cheng2020panoptic}
Bowen Cheng, Maxwell~D Collins, Yukun Zhu, Ting Liu, Thomas~S Huang, Hartwig
  Adam, and Liang-Chieh Chen.
\newblock Panoptic-deeplab: A simple, strong, and fast baseline for bottom-up
  panoptic segmentation.
\newblock In \emph{Proceedings of the IEEE/CVF conference on computer vision
  and pattern recognition}, pp.\  12475--12485, 2020{\natexlab{a}}.

\bibitem[Cheng et~al.(2021)Cheng, Girshick, Doll{\'a}r, Berg, and
  Kirillov]{cheng2021boundary}
Bowen Cheng, Ross Girshick, Piotr Doll{\'a}r, Alexander~C Berg, and Alexander
  Kirillov.
\newblock Boundary iou: Improving object-centric image segmentation evaluation.
\newblock In \emph{Proceedings of the IEEE/CVF Conference on Computer Vision
  and Pattern Recognition}, pp.\  15334--15342, 2021.

\bibitem[Cheng et~al.(2020{\natexlab{b}})Cheng, Chung, Tai, and
  Tang]{cheng2020cascadepsp}
Ho~Kei Cheng, Jihoon Chung, Yu-Wing Tai, and Chi-Keung Tang.
\newblock Cascadepsp: Toward class-agnostic and very high-resolution
  segmentation via global and local refinement.
\newblock In \emph{Proceedings of the IEEE/CVF Conference on Computer Vision
  and Pattern Recognition}, pp.\  8890--8899, 2020{\natexlab{b}}.

\bibitem[Cheng et~al.(2020{\natexlab{c}})Cheng, Wang, Huang, and
  Liu]{cheng2020boundary}
Tianheng Cheng, Xinggang Wang, Lichao Huang, and Wenyu Liu.
\newblock Boundary-preserving mask r-cnn.
\newblock In \emph{European conference on computer vision}, pp.\  660--676.
  Springer, 2020{\natexlab{c}}.

\bibitem[Cordts et~al.(2016)Cordts, Omran, Ramos, Rehfeld, Enzweiler, Benenson,
  Franke, Roth, and Schiele]{cordts2016Cityscapes}
Marius Cordts, Mohamed Omran, Sebastian Ramos, Timo Rehfeld, Markus Enzweiler,
  Rodrigo Benenson, Uwe Franke, Stefan Roth, and Bernt Schiele.
\newblock The cityscapes dataset for semantic urban scene understanding.
\newblock In \emph{Proceedings of the IEEE conference on computer vision and
  pattern recognition}, pp.\  3213--3223, 2016.

\bibitem[Dias \& Medeiros(2018)Dias and Medeiros]{dias2018semantic}
Philipe~Ambrozio Dias and Henry Medeiros.
\newblock Semantic segmentation refinement by monte carlo region growing of
  high confidence detections.
\newblock In \emph{Asian Conference on Computer Vision}, pp.\  131--146.
  Springer, 2018.

\bibitem[Dong et~al.(2021)Dong, Zeng, Wang, Zhang, and Wei]{dong2021solq}
Bin Dong, Fangao Zeng, Tiancai Wang, Xiangyu Zhang, and Yichen Wei.
\newblock Solq: Segmenting objects by learning queries.
\newblock \emph{Advances in Neural Information Processing Systems},
  34:\penalty0 21898--21909, 2021.

\bibitem[Gupta et~al.(2019)Gupta, Dollar, and Girshick]{gupta2019lvis}
Agrim Gupta, Piotr Dollar, and Ross Girshick.
\newblock Lvis: A dataset for large vocabulary instance segmentation.
\newblock In \emph{Proceedings of the IEEE/CVF conference on computer vision
  and pattern recognition}, pp.\  5356--5364, 2019.

\bibitem[He et~al.(2017)He, Gkioxari, Doll{\'a}r, and Girshick]{he2017mask}
Kaiming He, Georgia Gkioxari, Piotr Doll{\'a}r, and Ross Girshick.
\newblock Mask r-cnn.
\newblock In \emph{Proceedings of the IEEE international conference on computer
  vision}, pp.\  2961--2969, 2017.

\bibitem[Huang et~al.(2019)Huang, Huang, Gong, Huang, and Wang]{huang2019mask}
Zhaojin Huang, Lichao Huang, Yongchao Gong, Chang Huang, and Xinggang Wang.
\newblock Mask scoring r-cnn.
\newblock In \emph{Proceedings of the IEEE/CVF conference on computer vision
  and pattern recognition}, pp.\  6409--6418, 2019.

\bibitem[Kang et~al.(2020)Kang, Lee, Park, Ryu, and Kim]{kang2020bshapenet}
Ba~Rom Kang, Hyunku Lee, Keunju Park, Hyunsurk Ryu, and Ha~Young Kim.
\newblock Bshapenet: Object detection and instance segmentation with bounding
  shape masks.
\newblock \emph{Pattern Recognition Letters}, 131:\penalty0 449--455, 2020.

\bibitem[Ke et~al.(2021)Ke, Tai, and Tang]{ke2021deep}
Lei Ke, Yu-Wing Tai, and Chi-Keung Tang.
\newblock Deep occlusion-aware instance segmentation with overlapping bilayers.
\newblock In \emph{Proceedings of the IEEE/CVF conference on computer vision
  and pattern recognition}, pp.\  4019--4028, 2021.

\bibitem[Ke et~al.(2022)Ke, Danelljan, Li, Tai, Tang, and Yu]{ke2022mask}
Lei Ke, Martin Danelljan, Xia Li, Yu-Wing Tai, Chi-Keung Tang, and Fisher Yu.
\newblock Mask transfiner for high-quality instance segmentation.
\newblock In \emph{Proceedings of the IEEE/CVF Conference on Computer Vision
  and Pattern Recognition}, pp.\  4412--4421, 2022.

\bibitem[Kirillov et~al.(2020)Kirillov, Wu, He, and
  Girshick]{kirillov2020pointrend}
Alexander Kirillov, Yuxin Wu, Kaiming He, and Ross Girshick.
\newblock Pointrend: Image segmentation as rendering.
\newblock In \emph{Proceedings of the IEEE/CVF conference on computer vision
  and pattern recognition}, pp.\  9799--9808, 2020.

\bibitem[Kr{\"a}henb{\"u}hl \& Koltun(2011)Kr{\"a}henb{\"u}hl and
  Koltun]{krahenbuhl2011efficient}
Philipp Kr{\"a}henb{\"u}hl and Vladlen Koltun.
\newblock Efficient inference in fully connected crfs with gaussian edge
  potentials.
\newblock \emph{Advances in neural information processing systems}, 24, 2011.

\bibitem[Li et~al.(2017)Li, Qi, Dai, Ji, and Wei]{li2017fully}
Yi~Li, Haozhi Qi, Jifeng Dai, Xiangyang Ji, and Yichen Wei.
\newblock Fully convolutional instance-aware semantic segmentation.
\newblock In \emph{Proceedings of the IEEE conference on computer vision and
  pattern recognition}, pp.\  2359--2367, 2017.

\bibitem[Lin et~al.(2014)Lin, Maire, Belongie, Hays, Perona, Ramanan,
  Doll{\'a}r, and Zitnick]{lin2014microsoft}
Tsung-Yi Lin, Michael Maire, Serge Belongie, James Hays, Pietro Perona, Deva
  Ramanan, Piotr Doll{\'a}r, and C~Lawrence Zitnick.
\newblock Microsoft coco: Common objects in context.
\newblock In \emph{European conference on computer vision}, pp.\  740--755.
  Springer, 2014.

\bibitem[Lin et~al.(2017)Lin, Doll{\'a}r, Girshick, He, Hariharan, and
  Belongie]{lin2017feature}
Tsung-Yi Lin, Piotr Doll{\'a}r, Ross Girshick, Kaiming He, Bharath Hariharan,
  and Serge Belongie.
\newblock Feature pyramid networks for object detection.
\newblock In \emph{Proceedings of the IEEE conference on computer vision and
  pattern recognition}, pp.\  2117--2125, 2017.

\bibitem[Long et~al.(2015)Long, Shelhamer, and Darrell]{long2015fully}
Jonathan Long, Evan Shelhamer, and Trevor Darrell.
\newblock Fully convolutional networks for semantic segmentation.
\newblock In \emph{Proceedings of the IEEE conference on computer vision and
  pattern recognition}, pp.\  3431--3440, 2015.

\bibitem[Ren et~al.(2015)Ren, He, Girshick, and Sun]{ren2015faster}
Shaoqing Ren, Kaiming He, Ross Girshick, and Jian Sun.
\newblock Faster r-cnn: Towards real-time object detection with region proposal
  networks.
\newblock \emph{Advances in neural information processing systems}, 28, 2015.

\bibitem[Shen et~al.(2017)Shen, Gan, Yan, and Zeng]{shen2017semantic}
Falong Shen, Rui Gan, Shuicheng Yan, and Gang Zeng.
\newblock Semantic segmentation via structured patch prediction, context crf
  and guidance crf.
\newblock In \emph{Proceedings of the IEEE Conference on Computer Vision and
  Pattern Recognition}, pp.\  1953--1961, 2017.

\bibitem[Shen et~al.(2021)Shen, Yang, Wei, Deng, Huang, Hua, Cheng, and
  Liang]{shen2021dct}
Xing Shen, Jirui Yang, Chunbo Wei, Bing Deng, Jianqiang Huang, Xian-Sheng Hua,
  Xiaoliang Cheng, and Kewei Liang.
\newblock Dct-mask: Discrete cosine transform mask representation for instance
  segmentation.
\newblock In \emph{Proceedings of the IEEE/CVF Conference on Computer Vision
  and Pattern Recognition}, pp.\  8720--8729, 2021.

\bibitem[Tang et~al.(2021)Tang, Chen, Li, Li, Zhang, and Hu]{tang2021look}
Chufeng Tang, Hang Chen, Xiao Li, Jianmin Li, Zhaoxiang Zhang, and Xiaolin Hu.
\newblock Look closer to segment better: Boundary patch refinement for instance
  segmentation.
\newblock \emph{arXiv preprint arXiv:2104.05239}, 2021.

\bibitem[Welling \& Kipf(2016)Welling and Kipf]{welling2016semi}
Max Welling and Thomas~N Kipf.
\newblock Semi-supervised classification with graph convolutional networks.
\newblock In \emph{J. International Conference on Learning Representations
  (ICLR 2017)}, 2016.

\bibitem[Wu et~al.(2019)Wu, Kirillov, Massa, Lo, and
  Girshick]{wu2019detectron2}
Yuxin Wu, Alexander Kirillov, Francisco Massa, Wan-Yen Lo, and Ross Girshick.
\newblock Detectron2.
\newblock \url{https://github.com/facebookresearch/detectron2}, 2019.

\bibitem[Yang \& Yan(2022)Yang and Yan]{yang2022on}
Xue Yang and Junchi Yan.
\newblock On the arbitrary-oriented object detection: Classification based
  approaches revisited.
\newblock \emph{International Journal of Computer Vision}, 130\penalty0
  (5):\penalty0 1340--1365, 2022.

\bibitem[Yang et~al.(2019)Yang, Yang, Yan, Zhang, Zhang, Guo, Sun, and
  Fu]{yang2019scrdet}
Xue Yang, Jirui Yang, Junchi Yan, Yue Zhang, Tengfei Zhang, Zhi Guo, Xian Sun,
  and Kun Fu.
\newblock Scrdet: Towards more robust detection for small, cluttered and
  rotated objects.
\newblock In \emph{Proceedings of the IEEE International Conference on Computer
  Vision}, pp.\  8232--8241, 2019.

\bibitem[Yang et~al.(2021{\natexlab{a}})Yang, Yan, Feng, and He]{yang2021r3det}
Xue Yang, Junchi Yan, Ziming Feng, and Tao He.
\newblock R3det: Refined single-stage detector with feature refinement for
  rotating object.
\newblock In \emph{Proceedings of the AAAI Conference on Artificial
  Intelligence}, volume~35, pp.\  3163--3171, 2021{\natexlab{a}}.

\bibitem[Yang et~al.(2021{\natexlab{b}})Yang, Yan, Ming, Wang, Zhang, and
  Tian]{yang2021rethinking}
Xue Yang, Junchi Yan, Qi~Ming, Wentao Wang, Xiaopeng Zhang, and Qi~Tian.
\newblock Rethinking rotated object detection with gaussian wasserstein
  distance loss.
\newblock In \emph{International Conference on Machine Learning}, pp.\
  11830--11841. PMLR, 2021{\natexlab{b}}.

\bibitem[Yang et~al.(2021{\natexlab{c}})Yang, Yang, Yang, Ming, Wang, Tian, and
  Yan]{yang2021learning}
Xue Yang, Xiaojiang Yang, Jirui Yang, Qi~Ming, Wentao Wang, Qi~Tian, and Junchi
  Yan.
\newblock Learning high-precision bounding box for rotated object detection via
  kullback-leibler divergence.
\newblock \emph{Advances in Neural Information Processing Systems},
  34:\penalty0 18381--18394, 2021{\natexlab{c}}.

\bibitem[Yang et~al.(2022)Yang, Zhang, Yang, Zhou, Wang, Tang, He, and
  Yan]{yang2022detecting}
Xue Yang, Gefan Zhang, Xiaojiang Yang, Yue Zhou, Wentao Wang, Jin Tang, Tao He,
  and Junchi Yan.
\newblock Detecting rotated objects as gaussian distributions and its 3-d
  generalization.
\newblock \emph{IEEE Transactions on Pattern Analysis and Machine
  Intelligence}, 2022.

\bibitem[Yang et~al.(2023)Yang, Yan, Liao, Yang, Tang, and
  He]{yang2023scrdet++}
Xue Yang, Junchi Yan, Wenlong Liao, Xiaokang Yang, Jin Tang, and Tao He.
\newblock Scrdet++: Detecting small, cluttered and rotated objects via
  instance-level feature denoising and rotation loss smoothing.
\newblock \emph{IEEE Transactions on Pattern Analysis and Machine
  Intelligence}, 45\penalty0 (2):\penalty0 2384--2399, 2023.

\bibitem[Zhang et~al.(2021)Zhang, Lu, Tan, Li, Zhang, Li, and
  Hu]{zhang2021refinemask}
Gang Zhang, Xin Lu, Jingru Tan, Jianmin Li, Zhaoxiang Zhang, Quanquan Li, and
  Xiaolin Hu.
\newblock Refinemask: Towards high-quality instance segmentation with
  fine-grained features.
\newblock In \emph{Proceedings of the IEEE/CVF conference on computer vision
  and pattern recognition}, pp.\  6861--6869, 2021.

\bibitem[Zhang et~al.(2020)Zhang, Tian, Shen, You, and Yan]{zhang2020mask}
Rufeng Zhang, Zhi Tian, Chunhua Shen, Mingyu You, and Youliang Yan.
\newblock Mask encoding for single shot instance segmentation.
\newblock In \emph{Proceedings of the IEEE/CVF conference on computer vision
  and pattern recognition}, pp.\  10226--10235, 2020.

\bibitem[Zheng et~al.(2015)Zheng, Jayasumana, Romera-Paredes, Vineet, Su, Du,
  Huang, and Torr]{zheng2015conditional}
Shuai Zheng, Sadeep Jayasumana, Bernardino Romera-Paredes, Vibhav Vineet,
  Zhizhong Su, Dalong Du, Chang Huang, and Philip~HS Torr.
\newblock Conditional random fields as recurrent neural networks.
\newblock In \emph{Proceedings of the IEEE international conference on computer
  vision}, pp.\  1529--1537, 2015.

\bibitem[Zhou et~al.(2022)Zhou, Yang, Zhang, Wang, Liu, Hou, Jiang, Liu, Yan,
  Lyu, Zhang, and Chen]{zhou2022mmrotate}
Yue Zhou, Xue Yang, Gefan Zhang, Jiabao Wang, Yanyi Liu, Liping Hou, Xue Jiang,
  Xingzhao Liu, Junchi Yan, Chengqi Lyu, Wenwei Zhang, and Kai Chen.
\newblock Mmrotate: A rotated object detection benchmark using pytorch.
\newblock In \emph{Proceedings of the 30th ACM International Conference on
  Multimedia}, pp.\  7331–7334, 2022.

\end{thebibliography}
\bibliographystyle{iclr2023_conference}

\appendix
% \section{Appendix}
% \appendix
  % \renewcommand{\appendixname}{Appendix~\Alph{section}}

\section{More Qualitative Results}

\subsection{Two-stage DCT}
We visualize some outputs of two-stage DCT and compare them with DCT-Mask to demonstrate the disadvantages of simply combining DCT-Mask with multi-stage progress.

\begin{figure}[!tb]
    \centering
	\subfloat[]{
		\includegraphics[width=0.5\textwidth,height= 0.1\textheight]{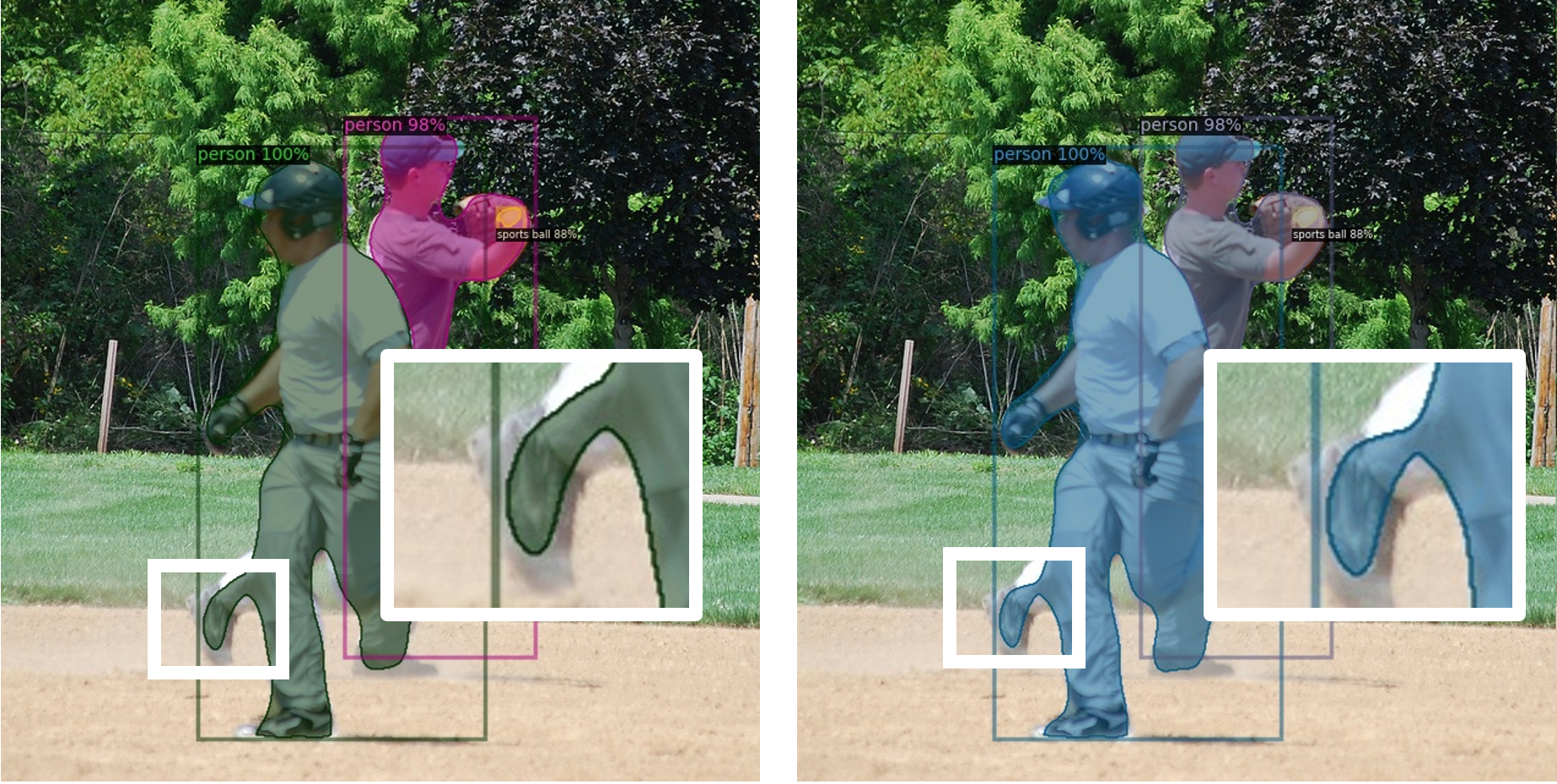}
	}
	\subfloat[]{
		\includegraphics[width=0.5\textwidth,height= 0.1\textheight]{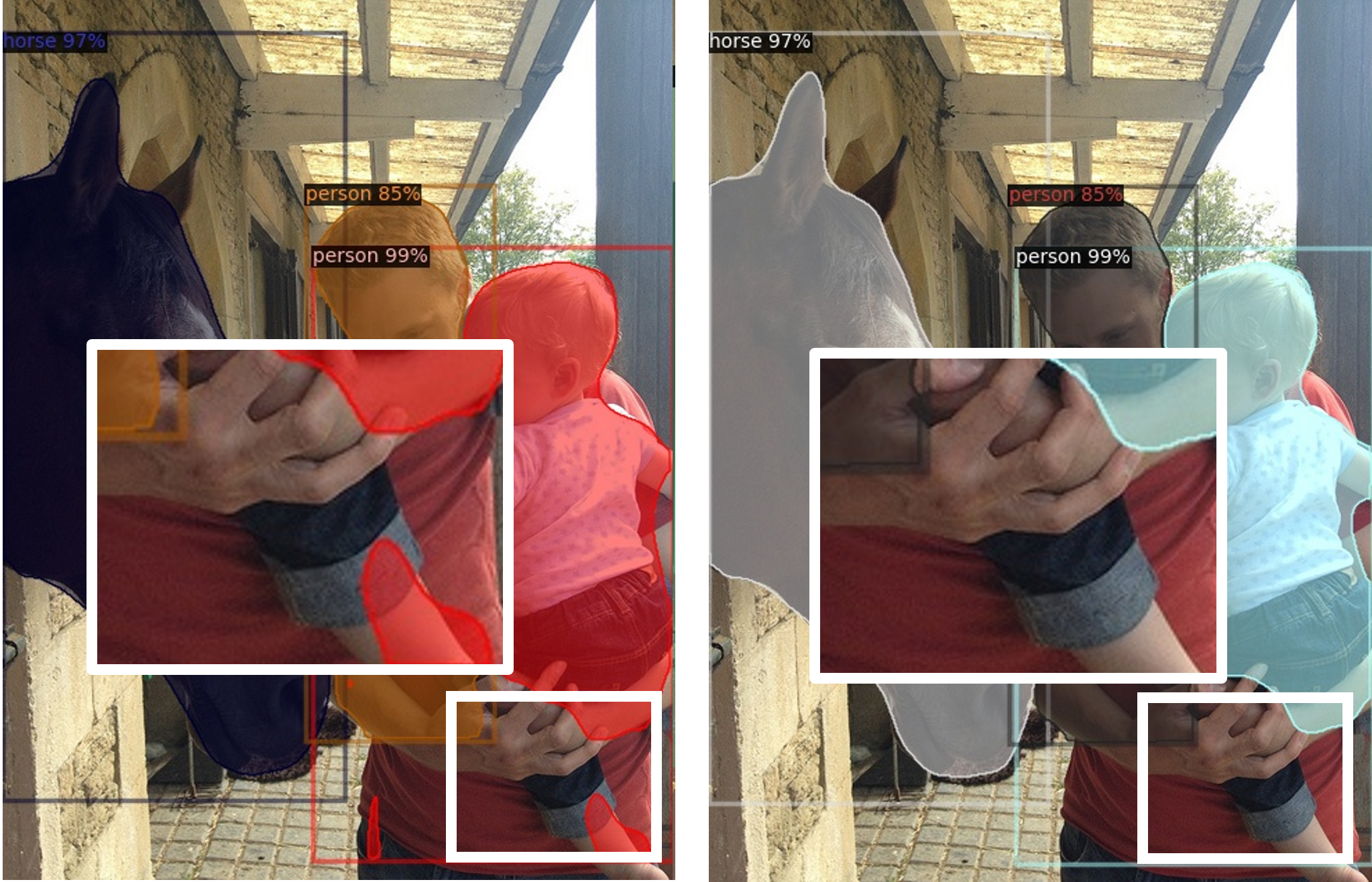}
	}
	\\
	\subfloat[]{
		\includegraphics[width=0.5\textwidth,height= 0.1\textheight]{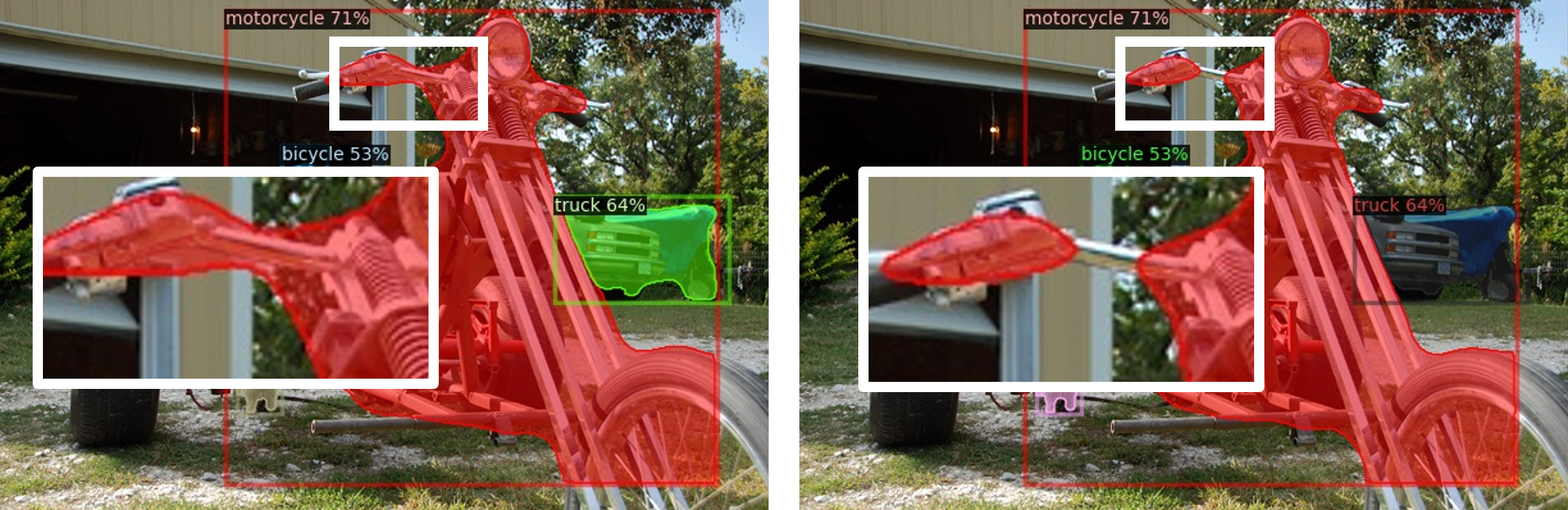}
	}
	\subfloat[]{
		\includegraphics[width=0.5\textwidth,height= 0.1\textheight]{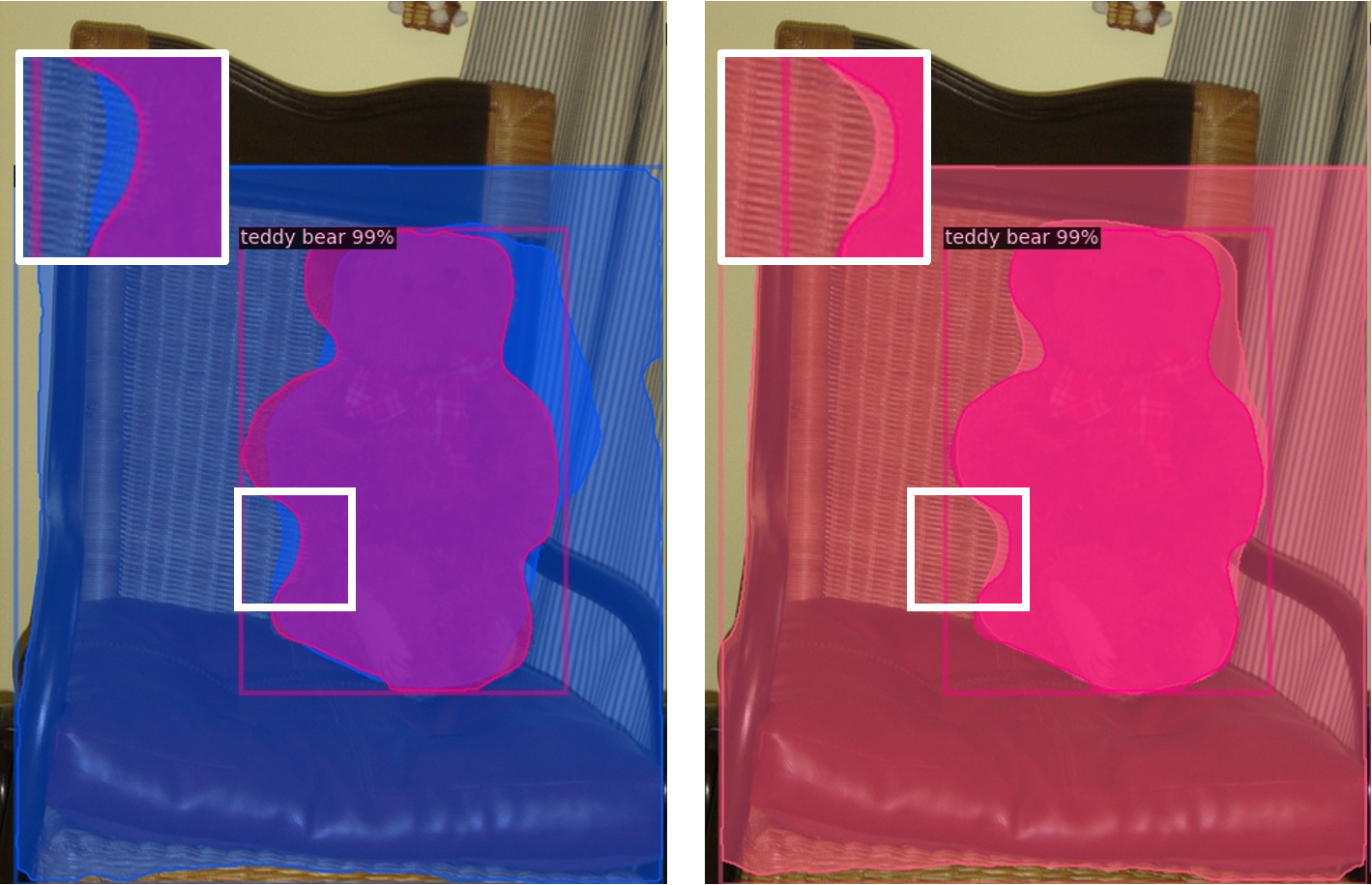}
	}
	% \vspace{2mm}
	%\vspace{-2mm}
 	\caption{Visualization of DCT-Mask (left) and two-stage DCT (right). Areas that were correctly predicted are influenced by the refinement.}
	\label{fig: influence_image}
	% \vspace{-2mm}
\end{figure}

As shown in Figure \ref{fig: influence_image}, in two-stage DCT, the areas that were previously correctly predicted may be influenced in refinement.
%, although the segmentation quality in some other regions is improved. 
The phenomenon further proves the difficulties in refining DCT vectors directly.

\subsection{Qualitative results on Cityscapes}
\begin{figure}[h]
	\centering
	%\subfloat[Ground Truth]{
		%\includegraphics[width=0.2\textheight]{figure/gt.png}%}
	\subfloat[Mask-RCNN]{
		\includegraphics[width=0.32\textwidth,height=0.3\textheight]{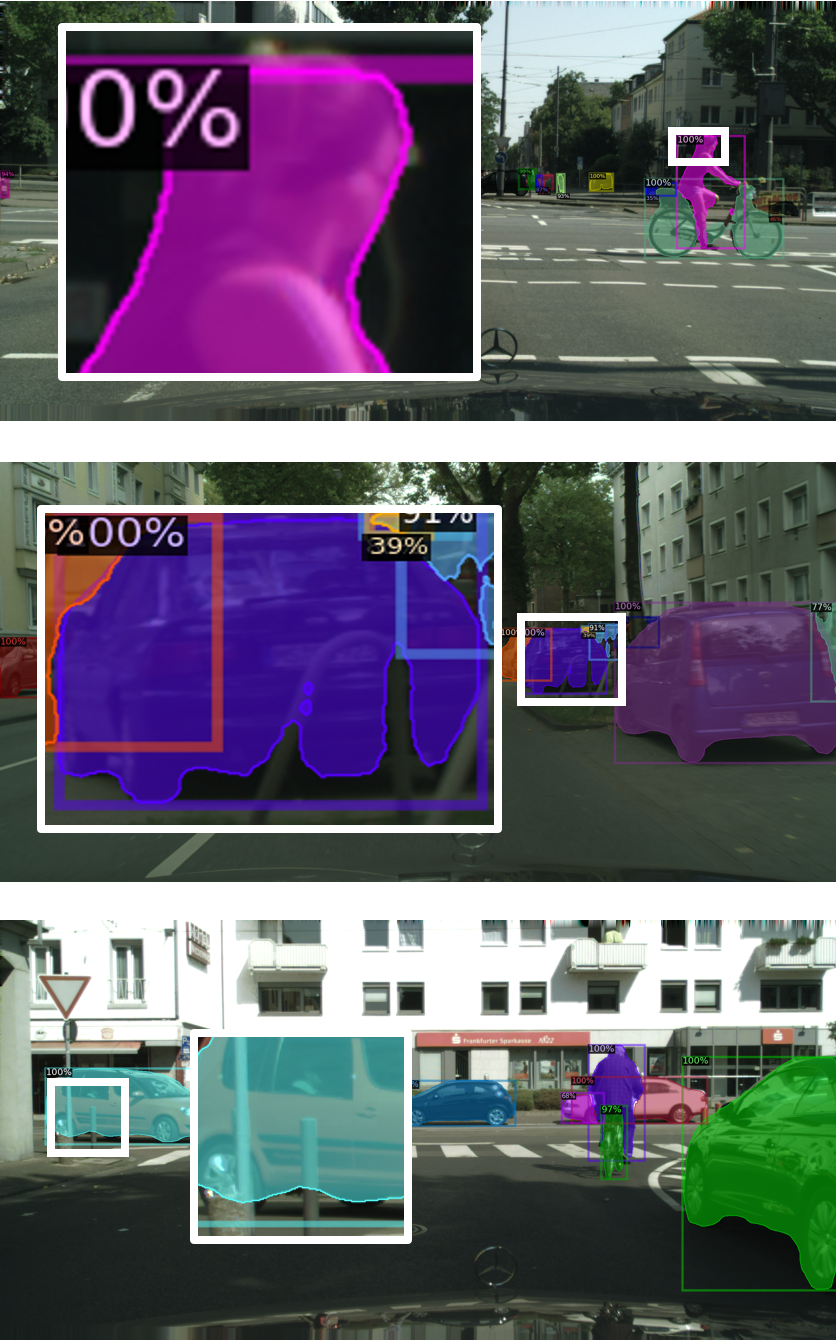}
	}
	\subfloat[DCT-Mask]{
		\includegraphics[width=0.33\textwidth,height=0.3\textheight]{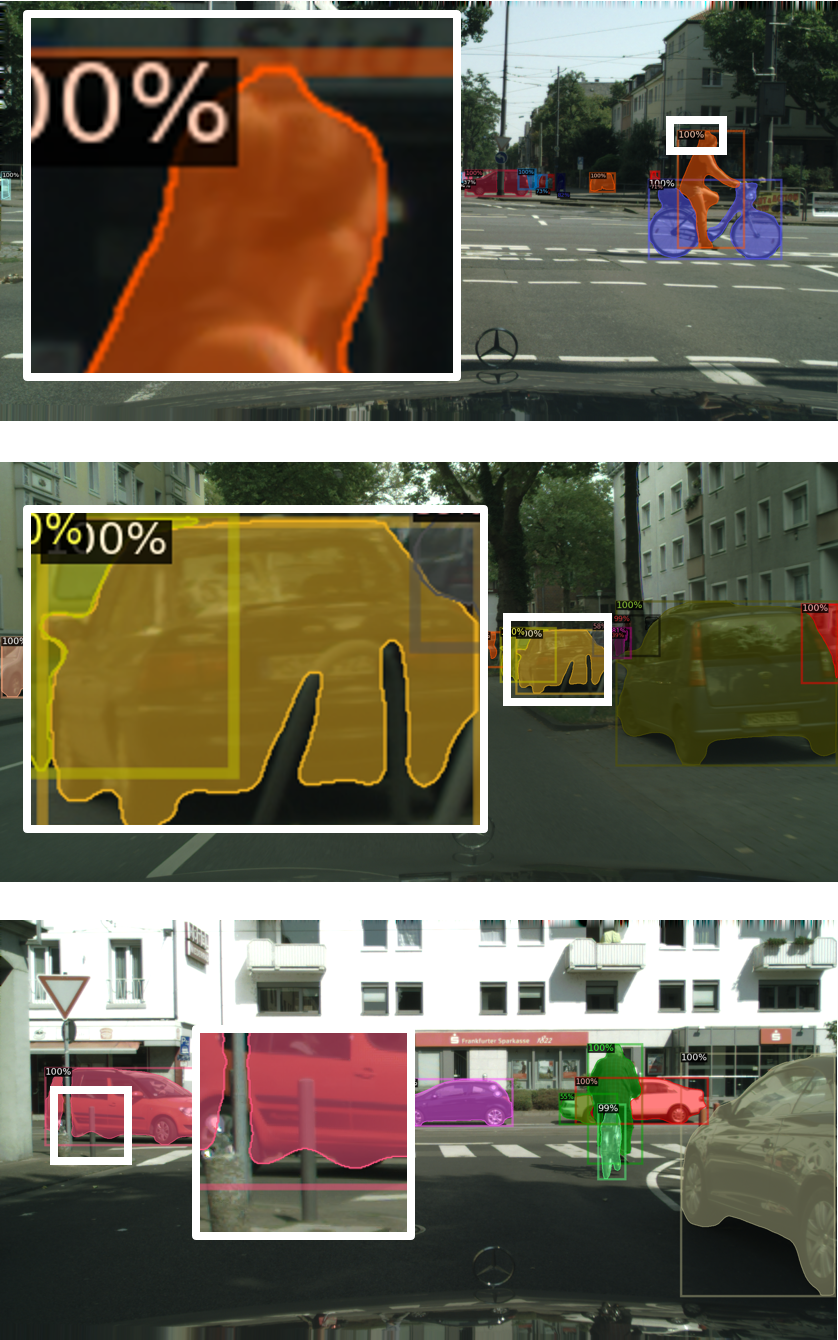}
	}
	%\subfloat[PatchDCT]{
		%\includegraphics[width=0.25\textwidth,height=0.45\textheight]{1p.png}
	%}
	\subfloat[PatchDCT]{
		\includegraphics[width=0.33\textwidth,height=0.3\textheight]{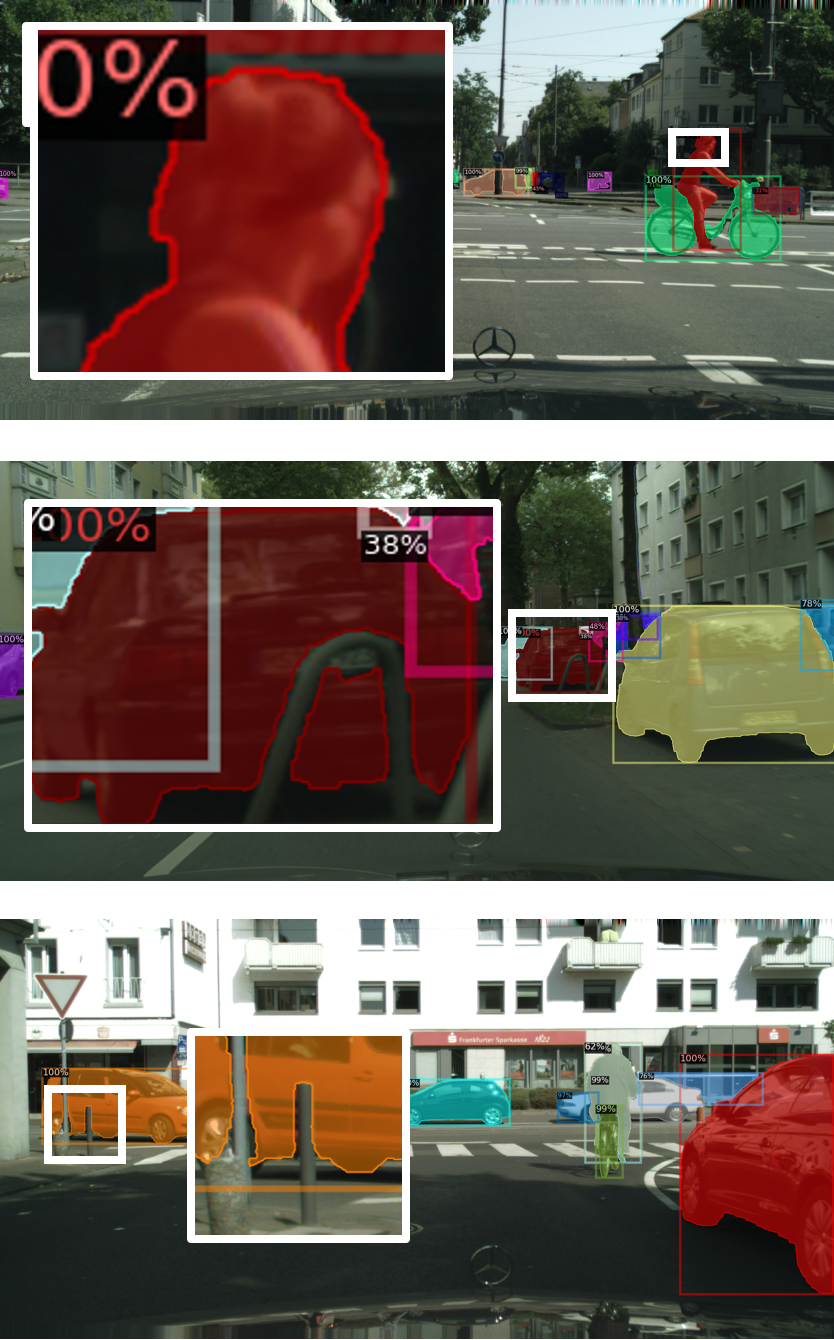}
	}
	\caption{Cityscapes example tuples from 
		%the ground truth masks(the first column), 
		Mask-RCNN, DCT-Mask, and PatchDCT. Mask-RCNN, DCT-Mask and PatchDCT are trained based on R50-FPN. PatchDCT generates masks with finer boundaries.}
	\label{fig:cityscapes_image}
 \vspace{-2mm}
\end{figure}

We show some qualitative results on Cityscapes in Figure \ref{fig:cityscapes_image}. In comparison with Mask-RCNN and DCT-Mask, PatchDCT generates finer boundaries that greatly improve the quality of masks.

\begin{figure}[!tb]
    \centering
	\subfloat[]{
		\includegraphics[width=0.95\textwidth,height=0.3\textheight]{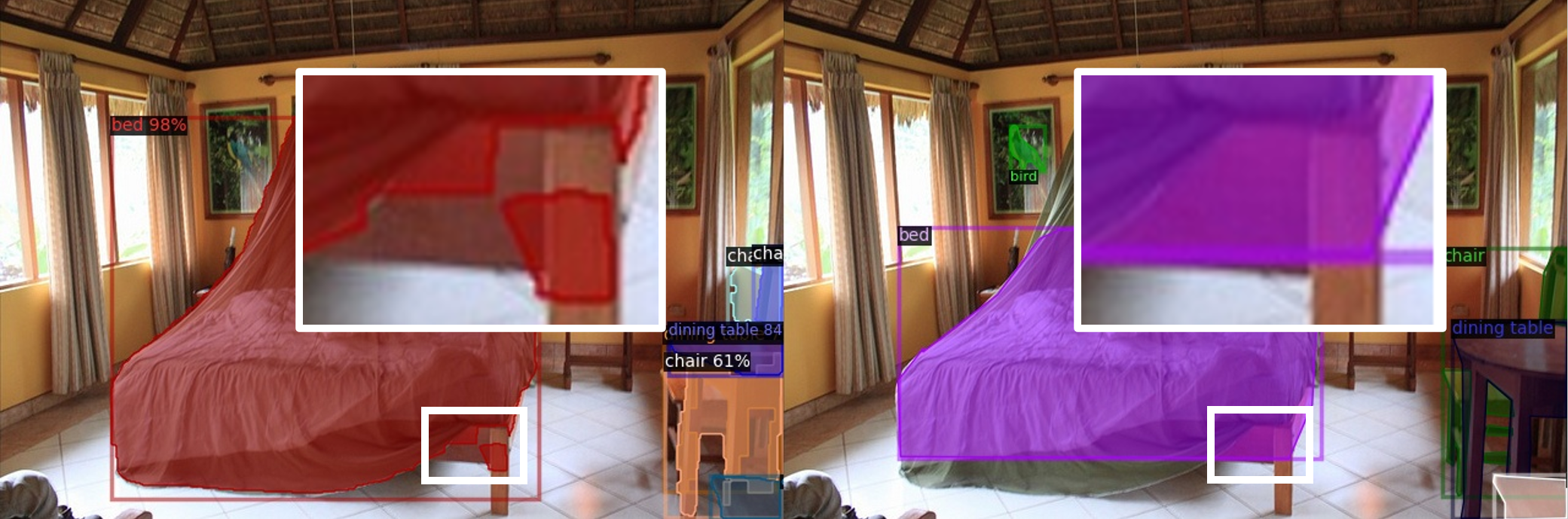}
	}\\
	\subfloat[]{
		\includegraphics[width=0.95\textwidth,height=0.3\textheight]{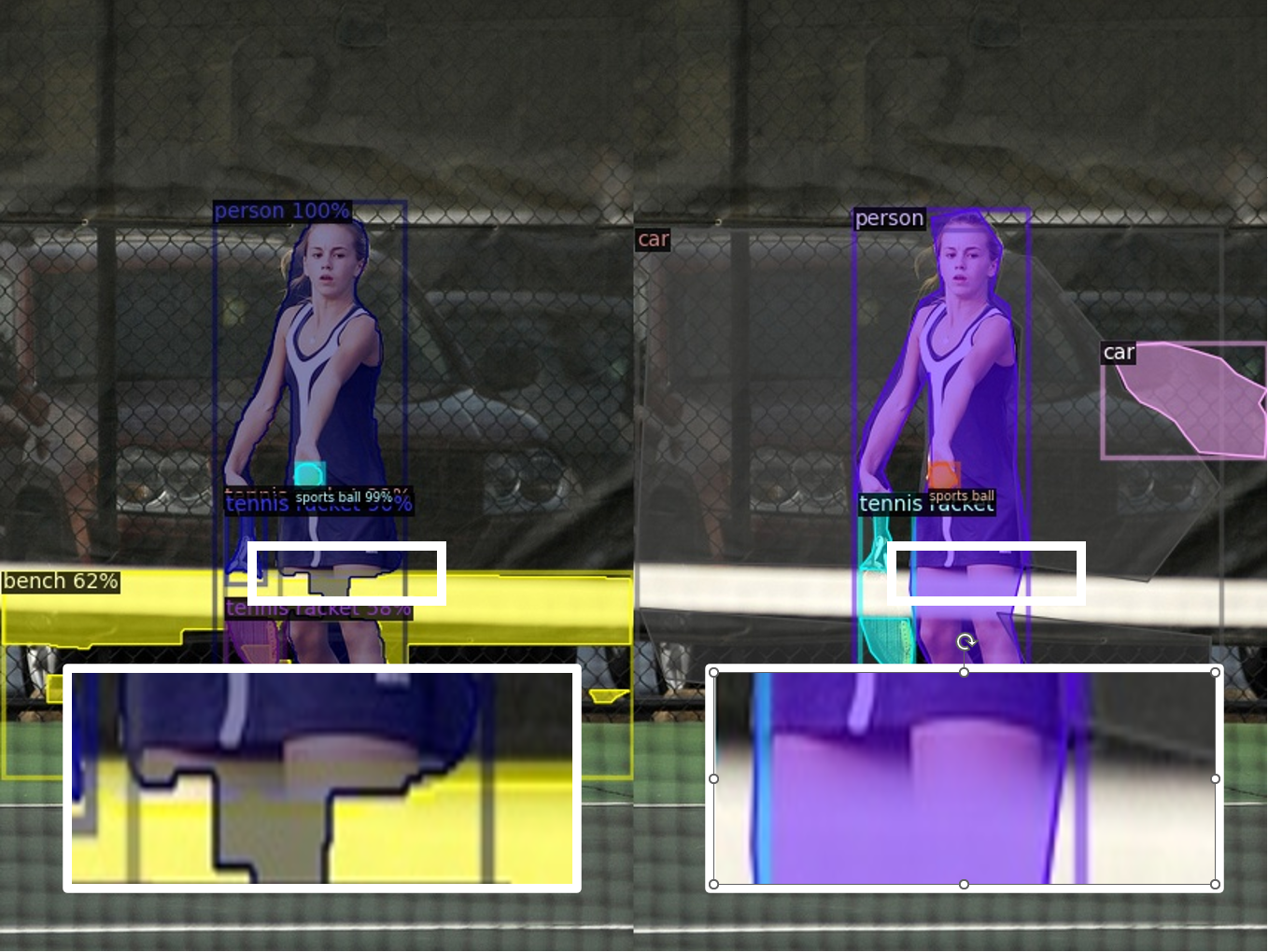}
	}\\
	\subfloat[]{
		\includegraphics[width=0.95\textwidth,height=0.3\textheight]{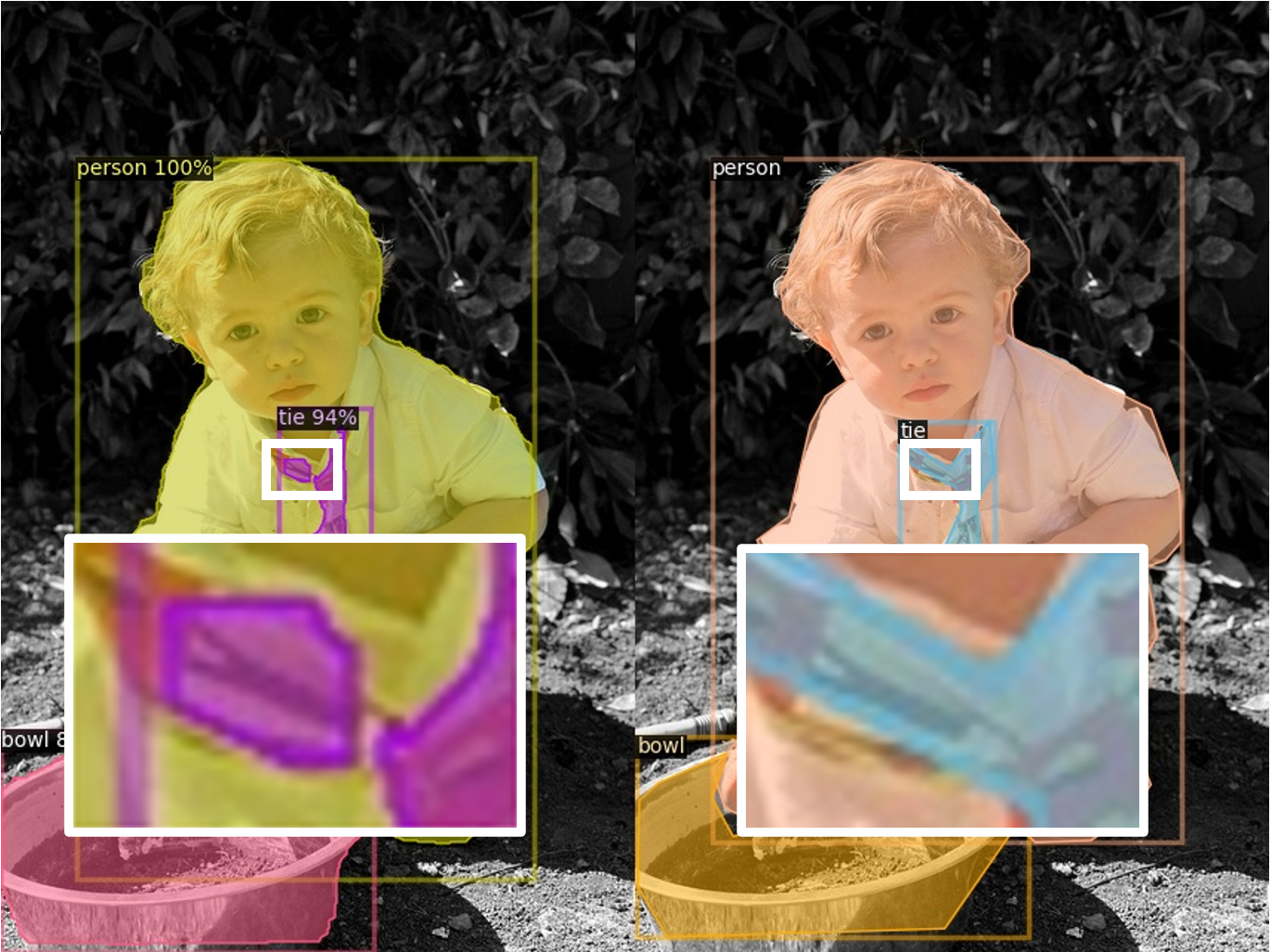}
	}\\
% 	\subfloat[]{
% 		\includegraphics[width=0.95\textwidth]{figure/b4.PNG}
% 	}
	% \vspace{2mm}
	%\vspace{-2mm}
 	\caption{Visualization of typical bad cases of our model, PatchDCT (left) and ground truth (right).}
	\label{fig: bad case}
	% \vspace{-2mm}
\end{figure}

\section{More Technical Details}
We prove that all elements except the DCCs for foreground patches are zero.
 
 It can be derived from Equation \ref{eq: dctcomp} that DCC is equal to the patch size $m$ in the foreground patch since $M_{m\times m}(x,y)=1$. 
 \begin{align}
	DCC = \frac{1}{m}\sum_{x=0}^{m-1}\sum_{y=0}^{m-1}
	M_{m\times m}(x,y) = m,
	\label{eq: dctcomp}
\end{align}

  Note that for a $m\times m$ patch $M^f_{m\times m}(u,v)$ Equation \ref{eq: dct} can be written as
\begin{align}
\small
	M^f_{m\times m}(u,v) = \frac{2}{m}C(u)C(v)\left(\sum_{x=0}^{m-1}A(x,u)\right)
	\left(\sum_{y=0}^{m-1}A(y,v)\right),
	%\label{eq: dct}
\end{align}
where $A(a,b)=cos\frac{(2a+1)b\pi}{2m}$. 

\noindent
If $u$ is odd,
\begin{align}
	A(m-1-x,u)&=cos\frac{(2(m-1-x)+1)u\pi}{2m}
	\notag
	\\&=cos\left(-\frac{(2x+1)u\pi}{2m}+u\pi\right)
	\notag
	\\&=-A(x,u),
\end{align}

\noindent
If $u$ is even and larger than zero, since from Euler's formula
\begin{align}
	e^{i\theta}=cos\theta+isin\theta,
\end{align}
We  have
\begin{align}
	\sum_{x=0}^{m-1}A(x,u)=\sum_{x=0}^{m-1}cos\frac{(2x+1)u\pi}{2m}
	\notag
	\\=Re\left(\sum_{x=0}^{m-1}e^{\frac{(2x+1)u \pi i}{2m}}\right)
	\notag
	\\=Re\left(e^{\frac{u\pi i}{2m}}\frac{1-e^{u\pi i}}{1-e^{\frac{u\pi i}{m}}}\right)=0,
\end{align}
Since $u$ is even,
\begin{align}
    e^{u\pi i} = cos(u\pi) + isin(u\pi) = 1,
\end{align}
We obtain 
\begin{align}
	\sum_{x=0}^{m-1}A(x,u)=0,\quad \forall u\neq 0, 
\end{align}
Therefore for foreground patches
\begin{align}
	\begin{split}
		M_{m\times m}^f(i,j)= \left \{
		\begin{array}{ll}
			m,&i=0,j=0,\\
			0,& otherwise. 	
		\end{array}
		\right.	
	\end{split}
\end{align}
This illustrates except the DCCs, elements of DCT vectors of foreground patches 
are all zero. 
\section{\wqr{Limitations and Future Outlook}}
\wqr{
In the process of visualization, we observe that the model may generate masks with holes. These problems usually occur in semantical ambiguous areas, and rarely in the center of the mask where the semantic information is very clear. We demonstrate some typical bad cases in Figure \ref{fig: bad case}. In these cases, the model either misclassifies these patches or generates imprecise patch DCT vectors, resulting in disconnected masks. We leave better classification and regression vectors as future work. In addition, we also plan to carry out further verification in other more challenging areas, such as aerial images, medical images, etc.
Taking aerial images as an example, this field still focuses on the research of object detection \citep{yang2019scrdet,yang2021r3det,yang2021rethinking,yang2021learning,yang2023scrdet++}, especially oriented object detection \citep{yang2022on,zhou2022mmrotate,yang2022detecting}, which lacks the exploration of more precise positioning tasks, i.e instance segmentation.
}

\end{document}